\documentclass[letterpaper, 10 pt, conference]{ieeeconf}  

\IEEEoverridecommandlockouts                              

\usepackage{graphics} 
\usepackage{epsfig} 
\usepackage{times} 
\usepackage{amsmath} 
\usepackage{amssymb}  
\usepackage[caption=false]{subfig}
\let\labelindent\relax
\usepackage[shortlabels]{enumitem}
\usepackage{multirow}
\usepackage{booktabs}

\newcommand{\pmnn}{\textsc{PMNN}}
\newcommand{\sensortraces}{\boldsymbol{S}}
\newcommand{\sensortracesdeviation}{\boldsymbol{\Delta} \sensortraces}

\newcommand{\sensorexp}{\sensortraces_\text{expected}}
\newcommand{\sensoract}{\sensortraces_\text{actual}}

\newcommand{\phasevariable}{p}
\newcommand{\phasevelocity}{u}

\usepackage[bookmarks=true]{hyperref}

\usepackage[disable]{todonotes}

\title{\LARGE \bf Learning Sensor Feedback Models from Demonstrations via Phase-Modulated Neural Networks}
\author{Giovanni Sutanto$^{1,2}$, Zhe Su$^{1,2}$, Stefan Schaal$^{1,2}$ and Franziska Meier$^{1,3}$
  \thanks{$^{1}$Autonomous Motion Department, MPI-IS, T\"ubingen,
    Germany.}%
  \thanks{$^{2}$CLMC-Lab, University of Southern California, Los Angeles, USA.}%
  \thanks{$^{3}$RSE-Lab, University of Washington, Seattle,
    USA.}%
  \thanks{This research was supported in part by National Science
    Foundation grants IIS-1205249, IIS-1017134, EECS-0926052, the
    Office of Naval Research, the Okawa Foundation, and the
    Max-Planck-Society.}
}

\begin{document}

\maketitle
\thispagestyle{empty}
\pagestyle{empty}

\begin{abstract}
In order to robustly execute a task under environmental uncertainty, a robot needs to be able to reactively adapt 
to changes arising in its environment. The environment changes are usually reflected in 
deviation from expected sensory traces. 
These deviations in sensory traces can be used to drive the motion adaptation, and for this purpose, a feedback model is required. The feedback model maps the deviations in sensory traces to the motion plan adaptation.
In this paper, we develop a general data-driven framework for learning a feedback model from demonstrations. 
We utilize a variant of a radial basis function network structure --with movement phases as kernel centers-- which can generally be applied to represent any feedback models for movement primitives. 
To demonstrate the effectiveness of our framework, we 
test it on 
the task of scraping on a tilt board.
In this task, we are learning a reactive policy in the form of orientation adaptation, based on deviations of tactile sensor traces. As a proof of concept of our method, we provide evaluations on an anthropomorphic robot.

\end{abstract}


\section{Introduction}
\label{sec:introduction}
	The ability to handle unexpected sensor events is key to robustly executing manipulation tasks. Humans, for instance, can predict how it should \emph{feel} to pick up an object and correct a grasp if the actual experience deviates from this prediction. Phrased differently, humans can map errors in sensory space to corrections in action space. In order to endow our robots with this ability, two problems need to be tackled: First, the system needs to be able to predict what sensor measurements to expect. Second, it needs to learn how to map deviations from those predictions to changes in actions.

Learning what sensor measurements to expect at any moment in time, anywhere in the state space, is a challenging problem with no known viable solution.
However, associating sensor information with successful executions of motion primitives has been shown to be promising \cite{pastor_IROS_2011_ASM,  pastor2013dynamic}. When such sensor traces have been associated with a primitive, the robot can try to correct the primitive's nominal actions when the actual sensor readings deviate from what is expected. 

In order to do so, a feedback model that maps errors in sensor space to the corrective actions needs to be acquired.
\begin{figure}[t]
	\centering
    \includegraphics[width=0.85\columnwidth]{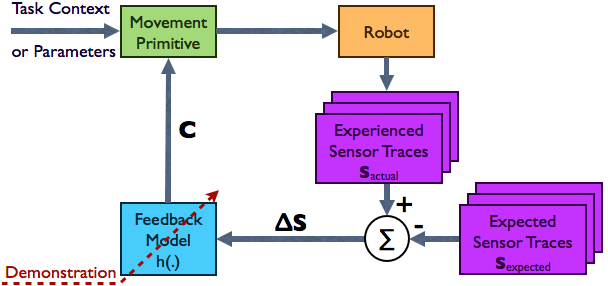}
    \vspace{-0.3cm}
 	\caption{\small{Proposed framework for learning behavior adaptation based on associative skill memories (ASMs).}}
 	\label{fig:BlockDiagramLearnFbTermsW_ASM}
    \vspace{-0.8cm}
\end{figure}
In initial implementations of such Associative Skill Memories (ASMs)  \cite{pastor_IROS_2011_ASM}, a linear feedback model was used. This feedback model essentially multiplies the sensor trace error with a manually defined feedback gain matrix to compute acceleration changes. While hand-designing feedback models can work well for specific problem settings, this approach is not expected to generalize beyond the scenario it was tuned for. Furthermore, when considering high-dimensional and multi-modal sensory input, such as haptic feedback, manually designing a feedback policy quickly becomes infeasible. For example, in this work we consider tactile-driven manipulation with tools. Manipulation tasks involving tools is challenging due to inaccurate tool kinematics models and non-rigid contacts between tactile sensors and the tool.

Thus, the larger goal of this research is to equip Associative Skill Memories with a general feedback modulation learning framework, as depicted in the block diagram in Figure \ref{fig:BlockDiagramLearnFbTermsW_ASM}. Data driven approaches to learning such feedback models have been proposed \cite{rai2014learning, icra2017_learning_feedback, Chebotar_IROS_2014} in the past. 
Here, we present a learning framework that improves such data-driven approaches in generality and experimental validation.
First we contribute towards the goal of generality by proposing the use of phase-modulated neural networks ({\pmnn}s). Our previous work \cite{icra2017_learning_feedback} shows that feedforward neural networks (FFNNs) have greater flexibility to learn feedback policies from human demonstrations than a hand-designed model. 
However, FFNNs cannot capture phase-dependent sensory features or corrective actions. Thus, in this paper, we introduce ({\pmnn}s), which can learn phase-dependent feedback models and show that this improves learning performance when compared to regular FFNNs. Second, we present detailed insight on our experimental pipeline for learning feedback models on a tactile-driven manipulation task. Furthermore, we extensively evaluate our learning approach on this manipulation task across multiple task variations and successfully deploy our approach on a real robot. 

This paper is organized as follows. Section~\ref{sec:background} provides some background on the motion primitive representation and related work. Section~\ref{sec:learn_fb_from_demo} presents the details of our approach for learning feedback models from demonstrations. We then present insights into our experimental setup in Section~\ref{sec:sys_desc}. Finally, we evaluate our approach in Section~\ref{sec:experiments} and conclude with Section~\ref{sec:discussion}.

\section{Background and Related Work}
\label{sec:background}
	Here we review background material on our chosen motion primitive representation and related work in learning feedback model approaches, including tactile feedback learning.
\subsection{Quaternion DMPs}\label{ssec:dmp}
The Associative Skill Memories framework, as proposed in \cite{pastor2013dynamic}, uses Dynamic Movement Primitives (DMPs) \cite{Ijspeert_NC_2013} as a motion primitive representation.
DMPs are a goal-directed behavior described as a set of differential equations with well-defined attractor dynamics. It is this formulation of DMPs as a set of differential equations that allows for online modulation from various inputs, such as sensor traces, in a manner that is conceptually straight forward and simple to implement, relative to other movement representations.

In our work, DMPs need to represent both position and orientation of the endeffector. 
We refer the reader to \cite{icra2017_learning_feedback} for our position DMP formulation. Here we focus on reviewing Quaternion DMPs, which we use for orientation representation in our learning-from-demonstration experiments. 

Quaternion DMPs were first introduced in \cite{pastor_IROS_2011_ASM}, and then improved in \cite{Kramberger_Humanoids16_GenOrientationDMP, Ude_ICRA14_OrientationDMP} to fully take into account the geometry of $\textit{SO}$(3). Like position DMPs, they consist of a \emph{transformation system} and a \emph{canonical system}, governing the evolution of the orientation state and movement phase, respectively. 

The \textit{transformation system} of a quaternion DMP is\footnote{For defining Quaternion DMPs, the operators $\circ$, ${}^{*}$ and the generalized log and exponential maps $\log(\cdot)$, and $\exp(\cdot)$ are required. The definition of these operators are stated in Equations~\ref{eq:QuatComposition}, \ref{eq:QuatConjugation}, \ref{eq:LogMapping}, and \ref{eq:ExpMapping} in the Appendix.}:
\begin{equation}
    \tau^2 \dot{\boldsymbol{\omega}} = \alpha_{\omega} \left( \beta_{\omega} 2 \log\left( \boldsymbol{Q}_{g} \circ \boldsymbol{Q}^{*} \right) - \tau \boldsymbol{\omega}\right) + \boldsymbol{f} + \boldsymbol{C}
    \label{eq:OriDMPTransformationSystem}
\end{equation}
where $\boldsymbol{Q}$ is a unit quaternion representing the orientation, $\boldsymbol{Q}_g$ is the goal orientation and $\boldsymbol{\omega}, \dot{\boldsymbol{\omega}}$ are the 3D angular velocity and angular acceleration, respectively. $\boldsymbol{f}$ and $\boldsymbol{C}$ are the 3D orientation \textit{forcing term} and \textit{feedback/coupling term}\footnote{Throughout this paper, we use the term \emph{feedback} and the term \emph{coupling term} interchangeably.}, respectively. 
The forcing term encodes the nominal behavior, while the coupling term encodes behavior adaptation which is commonly based on sensory feedback. In this paper, we focus on learning a feedback model that generates the coupling term, which is described in Sub-Section \ref{ssec:ct_background}. During unrolling, we integrate $\boldsymbol{Q}$ forward in time to generate the kinematic orientation trajectory as follows:
\begin{equation}
    \boldsymbol{Q}_{t+1} = \exp\left( \frac{\boldsymbol{\omega} \Delta t}{2} \right) \circ \boldsymbol{Q}_{t}
    \label{eq:QuatIntegration}
\end{equation}
where $\Delta t$ is the integration step size.
We set the constants $\alpha_{\omega} = 25$ and $\beta_{\omega} = \alpha_{\omega}/4$ to get a critically-damped system response when both forcing term and coupling term are zero. $\tau$ is set proportional to the motion duration.

The movement phase variable $\phasevariable$ and phase velocity $\phasevelocity$ are governed by the second-order \textit{canonical system} as follows:
\begin{equation}
	\tau \dot{\phasevelocity} = \alpha_{\phasevelocity} \left( \beta_{\phasevelocity} \left( 0 - \phasevariable \right) - \phasevelocity\right)
	\label{eq:2ndOrderCanonicalSystemP1}
\end{equation}
\begin{equation}
	\tau \dot{\phasevariable} = \phasevelocity
	\label{eq:2ndOrderCanonicalSystemP2}
\end{equation}
We set the constants $\alpha_{\phasevelocity} = 25$ and $\beta_{\phasevelocity} = \alpha_{\phasevelocity}/4$.
The phase variable $\phasevariable$ is initialized with 1 and will converge to 0. On the other hand, the phase velocity $\phasevelocity$ has initial value 0 and will converge to 0. 
Note, for a multi degree-of-freedom (DOF) system, each DOF has its own transformation system, but all DOFs share the same canonical system \cite{Ijspeert_NC_2013}. 
%

The \textit{forcing term} $\boldsymbol{f}$ governs the shape of the primitive and is represented as a weighted combination of $N$ basis functions $\psi_i$ with width parameter $h_i$ and center at $c_i$, as follows:
\begin{equation} 
  \boldsymbol{f}\left( \phasevariable, \phasevelocity; \boldsymbol{w} \right) = \frac{\sum_{i=1}^N \psi_i \left( \phasevariable \right) \boldsymbol{w}_i}{\sum_{j=1}^N \psi_j \left( \phasevariable \right)} \phasevelocity
  \label{eq:DMPForcingTerm}
\end{equation}
where
\begin{equation} 
  \psi_i \left( \phasevariable \right) =  \exp\left( -h_i \left( \phasevariable - c_i \right)^2 \right)
  \label{eq:GaussianBasisFunction}
\end{equation}
Note, because the forcing term $\boldsymbol{f}$ is modulated by the phase velocity $\phasevelocity$, it is initially $0$ and will converge back to $0$.

The $N$ basis function weights $\boldsymbol{w}_i$ in equation \ref{eq:DMPForcingTerm} are learned from human demonstrations of \textit{baseline/nominal behaviors}, by setting the target regression variable:
\begin{equation}
    \boldsymbol{f}_\text{target} = -\alpha_{\omega} (\beta_{\omega} 2 \log\left( \boldsymbol{Q}_{g,\text{bd}} \circ \boldsymbol{Q}_\text{bd}^{*} \right)
            - \tau \boldsymbol{\omega}_\text{bd}) + \tau^2 \dot{\boldsymbol{\omega}}_\text{bd}\nonumber
    \label{eq:OriDMPForcingTermExtraction}
\end{equation}
where \{$\boldsymbol{Q}_\text{bd}, \boldsymbol{\omega}_\text{bd}, \dot{\boldsymbol{\omega}}_\text{bd}$\} is the set of \textit{baseline/nominal} orientation behavior demonstrations. Then we can perform linear regression to identify parameters $\boldsymbol{w}$, as shown in \cite{Ijspeert_NC_2013}.

Finally, we include a \textit{goal evolution system} as follows:
\begin{equation}
    \tau \boldsymbol{\omega}_g = \alpha_{\omega_{g}} 2 \log \left( \boldsymbol{Q}_{G} \circ \boldsymbol{Q}_{g}^{*} \right) 
    \label{eq:OriDMPGoalEvolSystem}
\end{equation}
where $\boldsymbol{Q}_{g}$ and $\boldsymbol{Q}_{G}$ are the evolving and steady-state goal orientation, respectively. We set the constant $\alpha_{\omega_{g}} = \alpha_{\omega}/2$.
The goal evolution system has two important roles related to safety during the algorithm deployment on robot hardware. The first role, as mentioned in \cite{Ijspeert_NC_2013}, is to avoid discontinuous jumps in accelerations when the goal is suddenly moved. The second role, as mentioned in \cite{nemec2012action}, is to ensure continuity between the state at the end of one primitive and the state at the start of the next one when executing a sequence of primitives. Here we ensure continuity between primitives for both position and orientation DMPs by adopting \cite{nemec2012action}.

\subsection{Related Work on Learning Feedback Models}
The ability to adapt movement plans to changes in the environment requires feedback models. In previous work, researchers have hand-designed feedback models for specific purposes. For instance, \cite{Park_Humanoids_2008, Hoffmann_ICRA_2009} devised feedback models for obstacle avoidance. \cite{Khansari_IJRR_2016} designed a human-inspired feedback model for performing robotic surface-to-surface contact alignment based on force-torque sensing. Force-torque sensing is also used in \cite{pastor_IROS_2011_ASM}, where a hand-designed feedback gain matrix maps deviations from the expected force-torque measurements to the grasp plan adaptation.

Previous work on robotic tactile-driven manipulation with tools has tried to learn feedback models
to correct the position plans for handling uncertainty between tools and the environment, via reinforcement learning \cite{Chebotar_IROS_2014} or motor babbling \cite{hoffmann2014adaptive}. In our work, we propose to bootstrap the learning of feedback model from human demonstrations.

Abu-Dakka et al. \cite{Abu_Dakka2015} iteratively learned feedforward terms to improve a force-torque-guided task execution over trials, while fixing feedback models as constant gain matrices.

Learning by demonstrations is also employed in \cite{gams2015learning} to train separate feedback models for different environmental settings. Gaussian process regression is used to interpolate between these learned models to predict the required feedback model in a new environmental setting. Our work, directly uses a single model to handle multiple settings.

%
Kupcsik et al. \cite{Kupcsiketal15} learns the mapping from contexts --or environmental settings-- to DMP parameters. On the other hand, we learn the mapping from sensory input to the plan adaptation, abstracting the pre-specification of the context.

In \cite{Sung_ICRA_2017}, a partially-observable Markov decision process (POMDP), which is parameterized by deep recurrent neural networks, is used to represent a haptic feedback model. In general, POMDPs models are not explicitly provided with the information of the movement phase which is essential for making prediction on the next corrective action. Our proposed approach, can learn phase-dependent corrective actions.

\section{Learning Feedback Models via Phase-Modulated Neural Networks}
\label{sec:learn_fb_from_demo}
	\begin{figure}[ht]
    \vspace{-0.3cm}
	\centering
	\includegraphics[width=0.9\columnwidth]{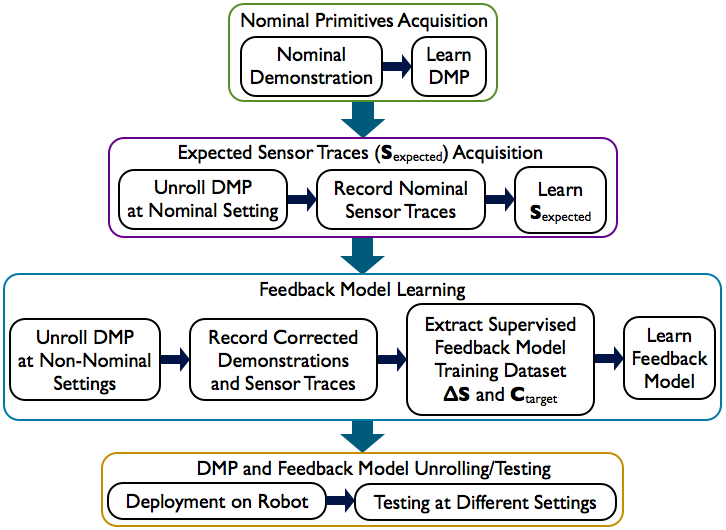}
    \vspace{-0.3cm}
 	\caption{Process pipeline of learning feedback model.}
	\label{fig:LearningFbTermsW_ASM_ProcessPipeline}
    \vspace{-0.5cm}
\end{figure}
In this section we describe our framework to learn general feedback models from human demonstrations. The process pipeline of learning feedback models is visualized in Figure \ref{fig:LearningFbTermsW_ASM_ProcessPipeline}. For a specific instance of this pipeline in our experiment, please refer to Sub-Section \ref{ssec:learning_pipeline_details}. Our framework comprises 3 core components: learning expected sensor traces; learning the feedback model to map sensor trace errors to corrections; and finally we introduce {\pmnn}s, a feedback model representation that is flexible enough to capture phase-dependent features and can learn across multiple task settings.
\subsection{Learning Expected Sensor Traces}
The core idea of ASMs \cite{pastor_IROS_2011_ASM}, \cite{pastor2013dynamic} rests on the insight that similar task executions should yield similar sensory events. Thus, an ASM of a task includes both a movement primitive as well as the expected sensor traces associated with this primitive's execution in a known environment. 
We term this execution as the primitive's \emph{nominal} behavior, the known environment as the nominal \emph{setting}, and the \emph{expected} sensor traces as $\sensorexp$. To learn the $\sensorexp$ model, we execute the nominal behavior and collect the experienced sensor measurements. Since these measurements are trajectories by nature, we can encode them using DMPs to become $\sensorexp$. 
This has the advantage that $\sensorexp$ is phase-aligned with the position and Quaternion DMP's execution, because they all share the same canonical system in Equations \ref{eq:2ndOrderCanonicalSystemP1} and \ref{eq:2ndOrderCanonicalSystemP2}.
%

\subsection{Learning Feedback Models from Demonstration}\label{ssec:ct_background}
When a movement primitive is executed under environment variations and/or uncertainties, the perceived sensor traces, denoted as \emph{actual} sensor traces $\sensoract$, tend to deviate from $\sensorexp$. The disparity $\sensoract - \sensorexp = \sensortracesdeviation$ can be used to drive corrections for adapting to the environmental changes causing the deviated sensor traces. Previous work \cite{Chebotar_IROS_2014, Kober_IROS_2008} uses reinforcement learning to learn these corrective behaviors, also in form of feedback models. However, learning a good feedback policy via trial-and-error from scratch is a very slow process. Therefore, we would like to bootstrap this process by learning feedback models from demonstrations. In our supervised learning framework, the disparity $\sensortracesdeviation$ is used as the input to a feedback model, mapping them to the motion plan adaptation or the coupling terms $\boldsymbol{C}$ (from Equation \ref{eq:OriDMPTransformationSystem}), 
as follows:
\begin{equation}
    \boldsymbol{C} = h(\sensoract - \sensorexp) = h(\sensortracesdeviation)
\end{equation}

We pose this as a regression problem, and similar to learning the \textit{nominal behavior}, we can also learn this feedback model $h$ from human demonstrations of \textit{corrected behavior}, i.e. the demonstrated behavior when the feedback is active. To perform the learning-from-demonstration, we need to extract the target output variable, i.e. the target coupling term $\boldsymbol{C}_\text{target}$, from demonstrations data, which can be done as follows:
\begin{equation}
    \boldsymbol{C}_\text{target} = -\alpha_{\omega} (\beta_{\omega} 2 \log\left( \boldsymbol{Q}_{g, cd} \circ \boldsymbol{Q}_{cd}^{*} \right) - \tau \boldsymbol{\omega}_{cd}) + \tau^2 \dot{\boldsymbol{\omega}}_{cd} - \boldsymbol{f}
    \label{eq:OriDMPCouplingTermExtraction}
\end{equation}
where \{$\boldsymbol{Q}_{cd}, \boldsymbol{\omega}_{cd}, \dot{\boldsymbol{\omega}}_{cd}$\} is the set of \textit{corrected} orientation behavior demonstration. 
Next, we describe our proposed general learning representation for the feedback model.

\subsection{Phase-Modulated Neural Network Structure}\label{ssec:pmnn}
\begin{figure}[h]
    \vspace{-0.2cm}
	\centering
	\includegraphics[width=0.95\columnwidth]{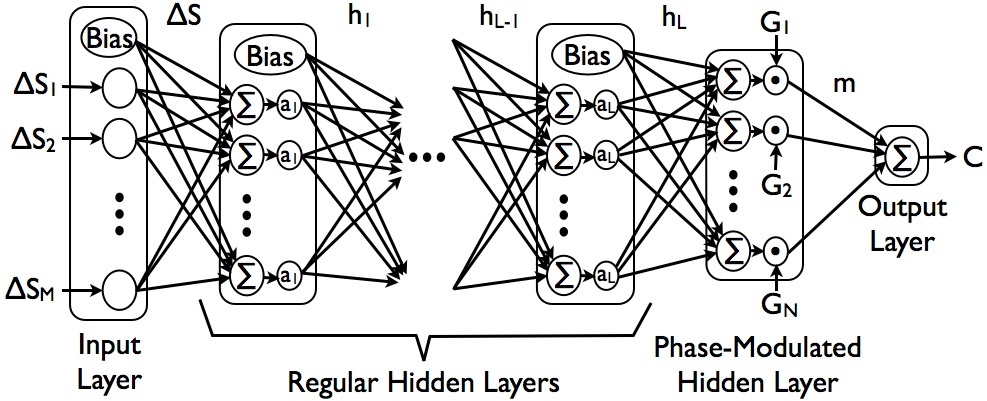}
    \vspace{-0.3cm}
 	\caption{Phase-modulated neural network ({\pmnn}) with one-dimensional output coupling term $C$.}
	\label{fig:special_design_neural_net}
    \vspace{-0.4cm}
\end{figure}

%
%
%
We use neural network (NN) structures for representing feedback term models due to its ability to learn task-relevant feature representations of high-dimensional inputs from data.
In this paper, we improve upon our previous work \cite{icra2017_learning_feedback}, in which we used a regular fully-connected feedforward neural network (FFNN) to represent the feedback model.
Our new neural network design is a variant of the radial basis function network (RBFN) \cite{Bishop_1991_RBFNN}, which we call the \textit{phase-modulated neural networks} ({\pmnn}s) as depicted in Figure \ref{fig:special_design_neural_net}.
{\pmnn} has an embedded structure that allows the encoding of a feedback model's dependency on the movement phase, which a FFNN structure lacks. We expect PMNN to model human adaptation better than FFNN because the same sensory deviation (NN input) may occur at different movement phases, but the form of the adaptation (NN output) will most likely be different.
There is also an alternative way of modeling phase-dependent adaptation behavior by using FFNN and including both phase variable $\phasevariable$ and phase velocity $\phasevelocity$ as inputs, together with the sensor trace deviations $\sensortracesdeviation$. However, there is no convergence guarantee on the adapted motion plan because the coupling term is not guaranteed to converge to zero, hence we may still need to hand-design an output post-processing similar to \cite{icra2017_learning_feedback} to ensure convergence. {\pmnn}, on the other hand, guarantees convergence due to the way we embed the information of phase velocity $\phasevelocity$ into the structure.

The {\pmnn} consists of:
\begin{itemize}
	\item \textit{input layer}\\
	    The input is $\sensortracesdeviation = \sensoract - \sensorexp$.
	\item \textit{regular hidden layers}\\
	    The regular hidden layers perform non-linear feature transformations on the high-dimensional inputs.
    	If there are $L$ layers, the output of $l$-th layer is:
    	\begin{equation}
    	h_{l} = \begin{cases}
    	\boldsymbol{a}_{l}\left( \boldsymbol{W}_{h_l \sensortracesdeviation} \sensortracesdeviation + \boldsymbol{b}_{h_l}  \right) &\text{for $l=1$}\\
    	\boldsymbol{a}_{l}\left( \boldsymbol{W}_{h_l h_{l-1}} \boldsymbol{h}_{l-1} + \boldsymbol{b}_{h_l}  \right) &\text{for $l=2,...,L$}
    	\end{cases}\nonumber
    	\end{equation}
    	$\boldsymbol{a}_{l}$ is the activation function of the $l$-th hidden layer, which can be \textrm{tanh}, \textsc{ReLU}, or others. $\boldsymbol{W}_{h_1 \sensortracesdeviation}$ is the weight matrix between the input layer and the first hidden layer. $\boldsymbol{W}_{h_l h_{l-1}}$ is the weight matrix between the $(l-1)$-th hidden layer and the $l$-th hidden layer. $\boldsymbol{b}_{h_l} $ is the bias vector at the $l$-th hidden layer.
	\item \textit{final hidden layer with phase kernel modulation}\\
    	This special and final hidden layer takes care of the dependency of the model on the movement phase.
    	The output of this layer is $\boldsymbol{m}$, which is defined as:
    	\begin{equation}
    	\boldsymbol{m} = \boldsymbol{G} \odot \left( \boldsymbol{W}_{m h_{L}} \boldsymbol{h}_{L} + \boldsymbol{b}_{m}  \right)
    	\label{eq:phase_modulation_layer}
    	\end{equation}
    	where $\odot$ denote element-wise product of vectors. $\boldsymbol{G} = \begin{bmatrix} 
    	G_1 & G_2 & \hdots & G_N
        \end{bmatrix}^T$ is the phase kernel modulation vector, and each component $G_i$ is defined as:
    	\begin{equation} 
            G_{i}\left( \phasevariable, \phasevelocity \right) = \frac{\psi_i \left( \phasevariable \right)}{\sum_{j=1}^N \psi_j \left( \phasevariable \right)} \phasevelocity \qquad i = 1,...,N
          	\label{eq:CouplingTermPhaseModulator}
        \end{equation}
        with phase variable $\phasevariable$ and phase velocity $\phasevelocity$, which comes from the second-order canonical system defined in Equation \ref{eq:2ndOrderCanonicalSystemP1} and \ref{eq:2ndOrderCanonicalSystemP2}. $\psi_i \left( \phasevariable \right)$ is the radial basis function (RBF) as defined in Equation \ref{eq:GaussianBasisFunction}.
        We use $N=25$ phase RBF kernels both in the {\pmnn}s as well as in the DMPs representation. The phase kernel centers have equal spacing in time, and we place these centers in the same way in the DMPs as well as in the {\pmnn}s.
	\item \textit{output layer}\\
    	The output is the one-dimensional coupling term $C$:
    	\begin{equation}
    	    C = \boldsymbol{w}_{C m}^{T} \boldsymbol{m}
    	\end{equation}
    	$\boldsymbol{w}_{C m}$ is the weight vector. Please note that there is no bias introduced in the output layer, and hence if $\boldsymbol{m} = \boldsymbol{0}$ --which occurs when the phase velocity $\phasevelocity$ is zero-- then $C$ is also zero. This ensures that $C$ is initially zero when a primitive is started. $C$ will also converge to zero because the phase velocity $\phasevelocity$ is converging to zero. This ensures the convergence of the adapted motion plan.
\end{itemize}
For an $M$-dimensional coupling term, we use $M$ separate {\pmnn}s with the same input vector $\sensortracesdeviation$ and the output of each {\pmnn} corresponds to each dimension of the coupling term. This separation 
allows each network to be optimized independently from each other.

We implemented {\pmnn} in TensorFlow \cite{TensorFlowBib}. To avoid overfitting, we used the dropout technique as introduced in \cite{Srivastava2014_NN_Dropouts}.

\section[Learning Tactile Feedback Models: System Overview and Experimental Setup]{Learning Tactile Feedback Models: \\System Overview and Experimental Setup}
\label{sec:sys_desc}
	\begin{figure}[ht]
  \vspace{-0.3cm}
  \centering
  \includegraphics[width=0.5\columnwidth]{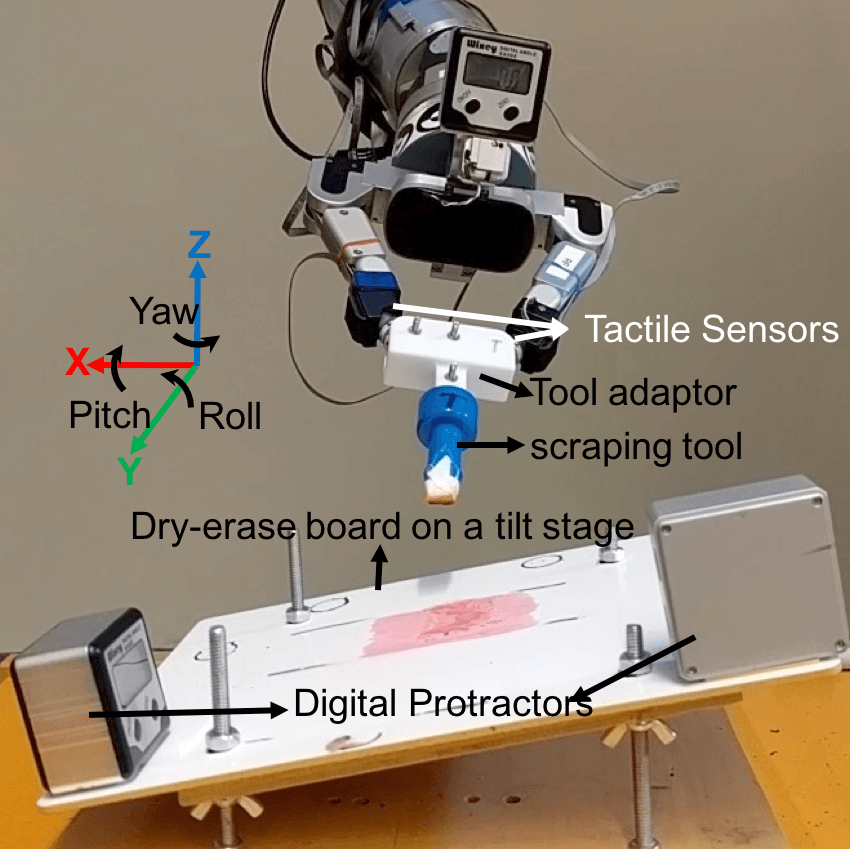}
  \vspace{-0.3cm}
  \caption{Experimental setup of the scraping task.}
  \label{fig:scraping_setup}
  \vspace{-0.5cm}
\end{figure}
This work is focused on learning to correct tactile-driven manipulation with tools. Our experimental scenario involves a demonstrator teaching our robot to perform a scraping task, utilizing a hand-held tool to scrape paint off the surface of a dry-erase board (see Figure~\ref{fig:scraping_setup}). The system is taught this skill at a nominal tilt angle, and needs to correct when the board is tilted away from that default angle. Neither vision nor motion capture system is used, thus we only rely on tactile sensing to inform the correction. 
One of the main challenges is that the tactile sensors interact indirectly with the board, i.e. through the tool adapter and the scraping tool via a non-rigid contact, and the robot does not explicitly encode the tool kinematics model. This makes hand-designing a feedback gain matrix difficult. Next, we explain the experimental setup and some lessons learned from the experiments.
\subsection{Hardware}
\label{ssec:hardware}
The demonstrations were performed on the right arm and the right hand of our bi-manual robot. The arm is a 7-degrees-of-freedom (DoF) Barrett WAM arm which is also equipped with a 6D force-torque (FT) sensor at the wrist.
The hand is a Barrett hand whose left and right fingers are equipped with biomimetic tactile sensors (BioTacs) \cite{wettels08}. The two BioTac-equipped fingers were setup to perform a pinch grasp on a tool adapter. The tool adapter is a 3D-printed object designed to hold a scraping tool with an 11mm-wide tool-tip.

The dry-erase board was mounted on a tilt stage whose orientation can be adjusted to create static tilts of $\pm 20^\circ$ in roll and/or pitch with respect to the robot global coordinates as shown in Figure \ref{fig:scraping_setup}. Two digital protractors with $0.1^\circ$ resolution (Wixey WR 300 Digital Angle Gauge) were used to measure the tilt angles during the experiment.
\subsection{Robot's Environmental Settings and Human Demonstrations with Sensory Traces Association}
\label{ssec:env_setting}
For our experiment, we considered 5 different \textit{settings}, and each setting is associated with a specific roll angle of the tilt stage, specifically at $0^\circ$, $2.5^\circ$, $5^\circ$, $7.5^\circ$, and $10^\circ$. At each setting, we fixed the pitch angle at $0^\circ$ and maintain the scraping path to be roughly at the same height. Hence, we assume that among the 6D pose action (x-y-z-pitch-roll-yaw), the necessary correction is only in the roll-orientation.
For each setting, we collected 15 demonstrations. The setting with roll angle at $0^\circ$ is selected as the \textit{nominal} setting, while the remaining settings become the \textit{corrected} ones.

For the demonstrated actions, we recorded the 6D pose trajectory of the right hand end-effector at 300 Hz rate, and along with these demonstrations, we also recorded the multi-dimensional sensory traces associated with this action. The sensory traces are the 38-dimensional tactile signals from the left and right BioTacs' electrodes, sampled at 100 Hz. 
\subsection{Learning Pipeline Details and Lessons Learned}
\label{ssec:learning_pipeline_details}
DMPs provide kinematic plans to be tracked with a position control scheme. However, for tactile-driven contact manipulation tasks such as the scraping task in this paper, using position control alone is not sufficient. In order to attain consistent tactile signals on task repetitions --during the demonstrations as well as during unrolling of the learned feedback models-- similar contact force profiles needs to be applied. Hence force control is required.

Moreover, while it is possible to perform corrected demonstrations solely by humans, the sensor traces obtained might be significantly different from the traces obtained during the robot's execution of the motion plan. This is problematic, because during learning and during prediction phases of the feedback terms, the input to the feedback models are different. Hence, instead we try to let the robot execute the nominal plans, and only provide correction by manually adjusting the robot's execution at different settings as necessary.

Therefore, we use the force-torque (FT) sensor in the robot's right wrist for FT control, with two purposes: (1) to maintain tool-tip contact with the board, such that consistent tactile signals are obtained, and (2) to provide compliance, allowing the human demonstrator to perform corrective action demonstration as the robot executes the nominal behavior.

For simplicity, we set the force control set points in our experiment to be constant. We need to set the force control set point carefully: if the downward force (in the z-axis direction) for contact maintenance is too big, the friction will block the robot from being able to execute the corrections as commanded by the feedback model. 
We found that 1 Newton is a reasonable value for the downward force control set point.
Regarding the learning process pipeline as depicted in Figure \ref{fig:LearningFbTermsW_ASM_ProcessPipeline}, here we provide the details in our experiment:
\begin{enumerate}
\item \textit{Nominal primitives acquisition}: While the robot is operating in the gravity-compensation mode and the tilt stage is at $0^\circ$ roll angle, the human demonstrator guided the robot's hand to kinesthetically perform a scraping task, which can be divided into three stages, each of which corresponds to a movement primitive:
\begin{enumerate}[(a)]
    \item \textit{primitive 1}: starting from its home position above the board, go down (in the z-axis direction) until the scraping tool made contact with the scraping board's surface (no orientation correction at this stage),
    \item \textit{primitive 2}: correct the tool-tip orientation such that it made a full flat tool-tip contact with the surface,
    \item \textit{primitive 3}: go forward in the y-axis direction while scraping paint off the surface, applying orientation correction as necessary to maintain full flat tool-tip contact with the surface.
\end{enumerate}
We used Zero Velocity Crossing (ZVC) method \cite{fod2002automated} and local minima search refinement on the 
velocity signal in the z and y axes, to find segmentation points of primitives 1 and 3, respectively. The remaining part -- between the end of primitives 1 and the beginning of primitive 3 -- becomes primitive 2. We encode each of these primitives with position and orientation DMPs.

\begin{table}[ht]
\vspace{-0.3cm}
\centering
\begin{tabular}{|c|c|c|c|}
\hline
\multirow{2}{*}{} & \multicolumn{3}{c|}{Force-Torque Control Activation Schedule} \\ \cline{2-4}
                  & Primitive 1 & Primitive 2 & Primitive 3 \\ \hline
Step 2       & -   & z 1 N     & z 1 N         \\ \hline
Step 3       & -   & z 1 N, roll 0 Nm     & z 1 N, roll 0 Nm         \\ \hline
Step 4       & -   & z 1 N     & z 1 N         \\ \hline
\end{tabular}
\label{table:ft_control_activation_schedule_table}
\caption{Force-torque control schedule for steps 2-4.}
\label{tab:ft_control_schedule_table}
\vspace{-0.6cm}
\end{table}

For the following pipeline steps (2, 3, and 4), in reference to Table \ref{tab:ft_control_schedule_table}, which indicates what force-torque control mode being active at each primitive of these steps. "z 1 N" refers to the 1 Newton downward z-axis proportional-integral (PI) force control, for making sure that consistent tactile signals are obtained at repetitions of the task; this is important for learning and making correction predictions properly. "roll 0 Nm" refers to the roll-orientation PI torque control at 0 Newton-meter, for allowing corrective action demonstration.
\item \textit{Expected sensor traces acquisition}:
Still with the tilt stage at $0^\circ$ roll angle, we unroll the nominal primitives 15 times and record the tactile sensor traces. We encode each dimension of the 38-dimensional sensor traces as $\sensorexp$, using the standard DMP formulation.
\begin{figure*}[ht!]
    \centering
    \includegraphics[trim={3cm 0 4cm 0},clip,width=0.35\textwidth]{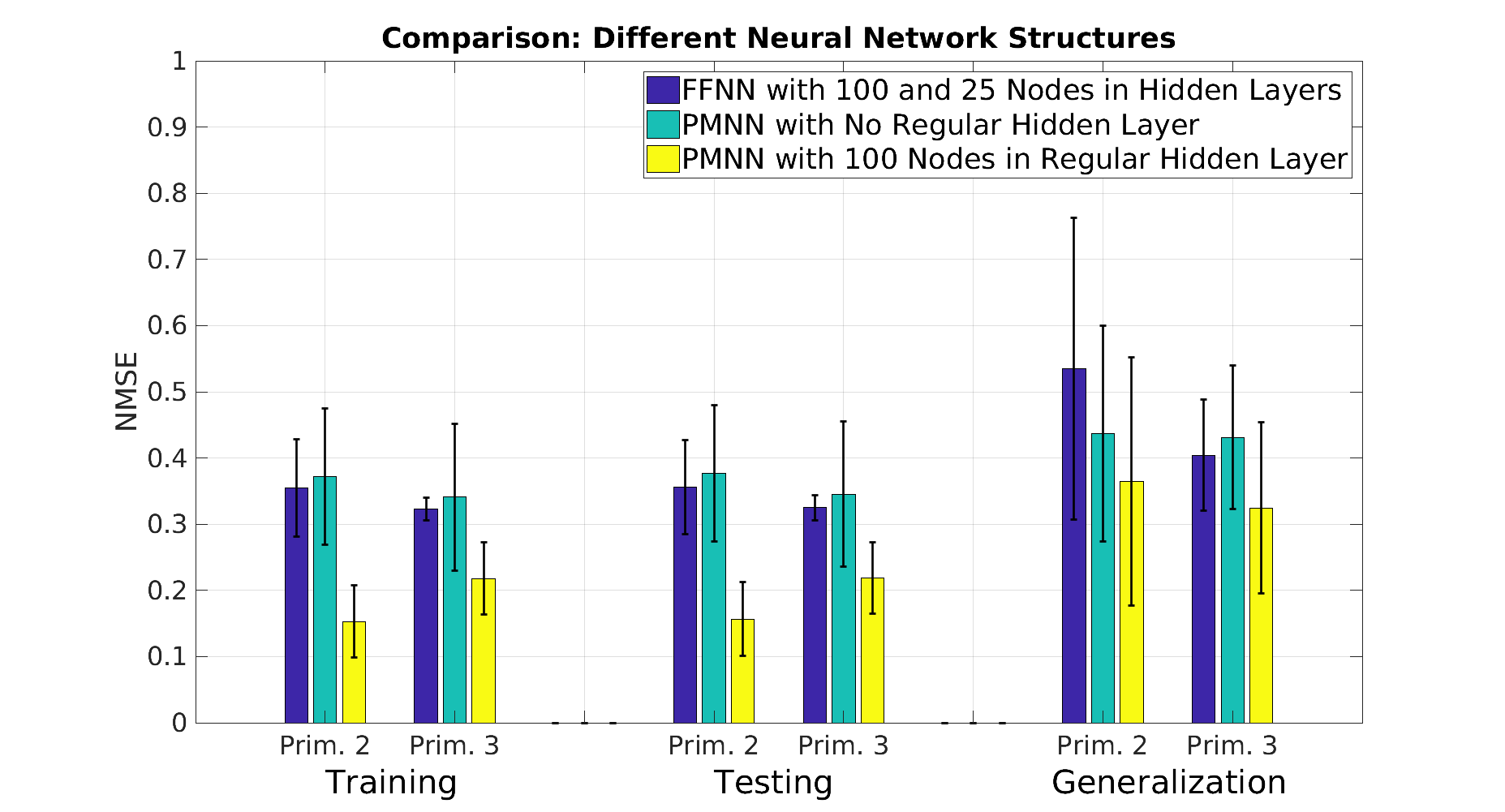}
    %
    \includegraphics[trim={3cm 0 4cm 0},clip,width=0.35\textwidth]{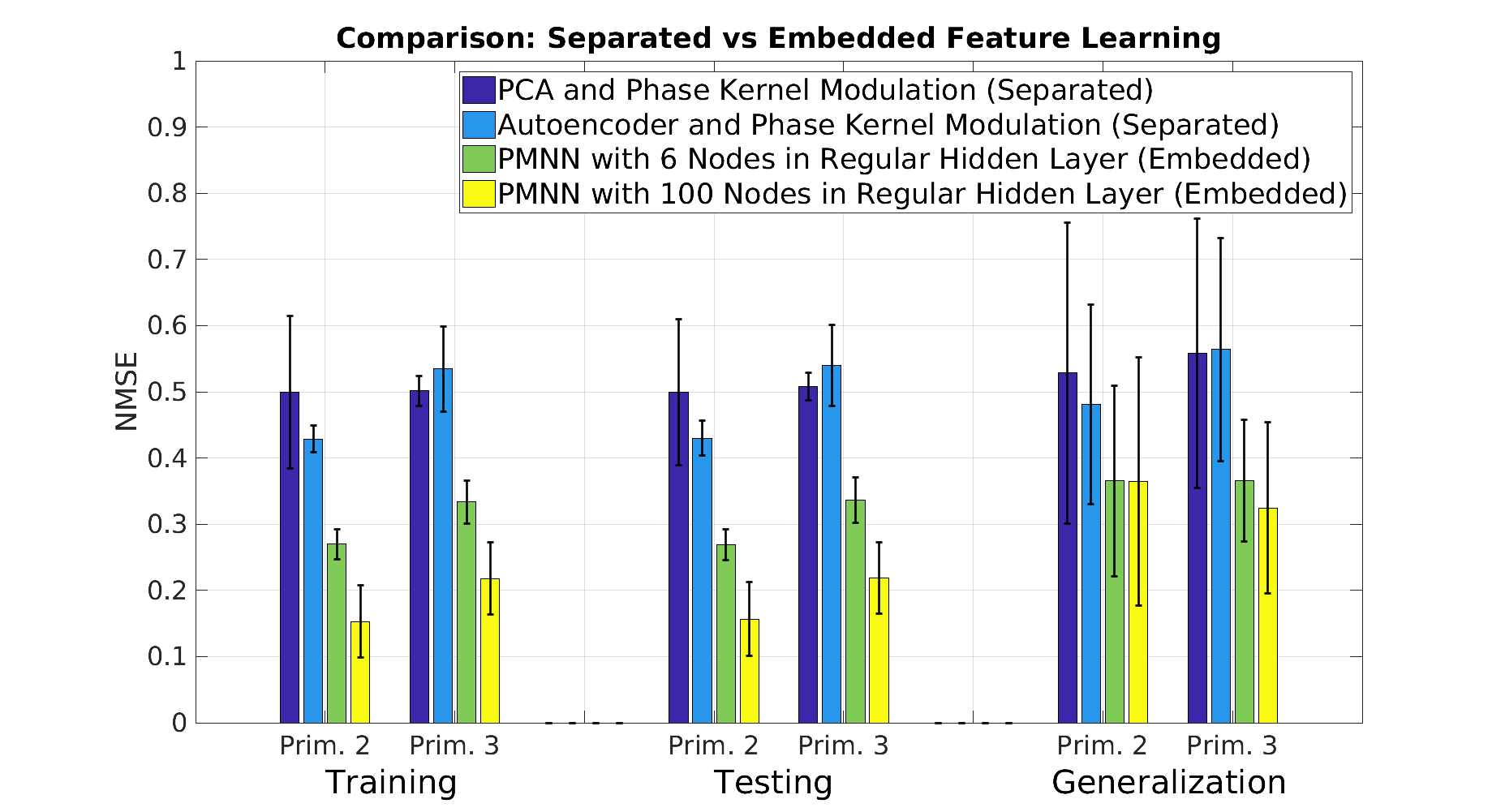}
    %
    \includegraphics[trim={3cm 0cm 3.5cm 1.95cm},clip,width=0.28\textwidth]{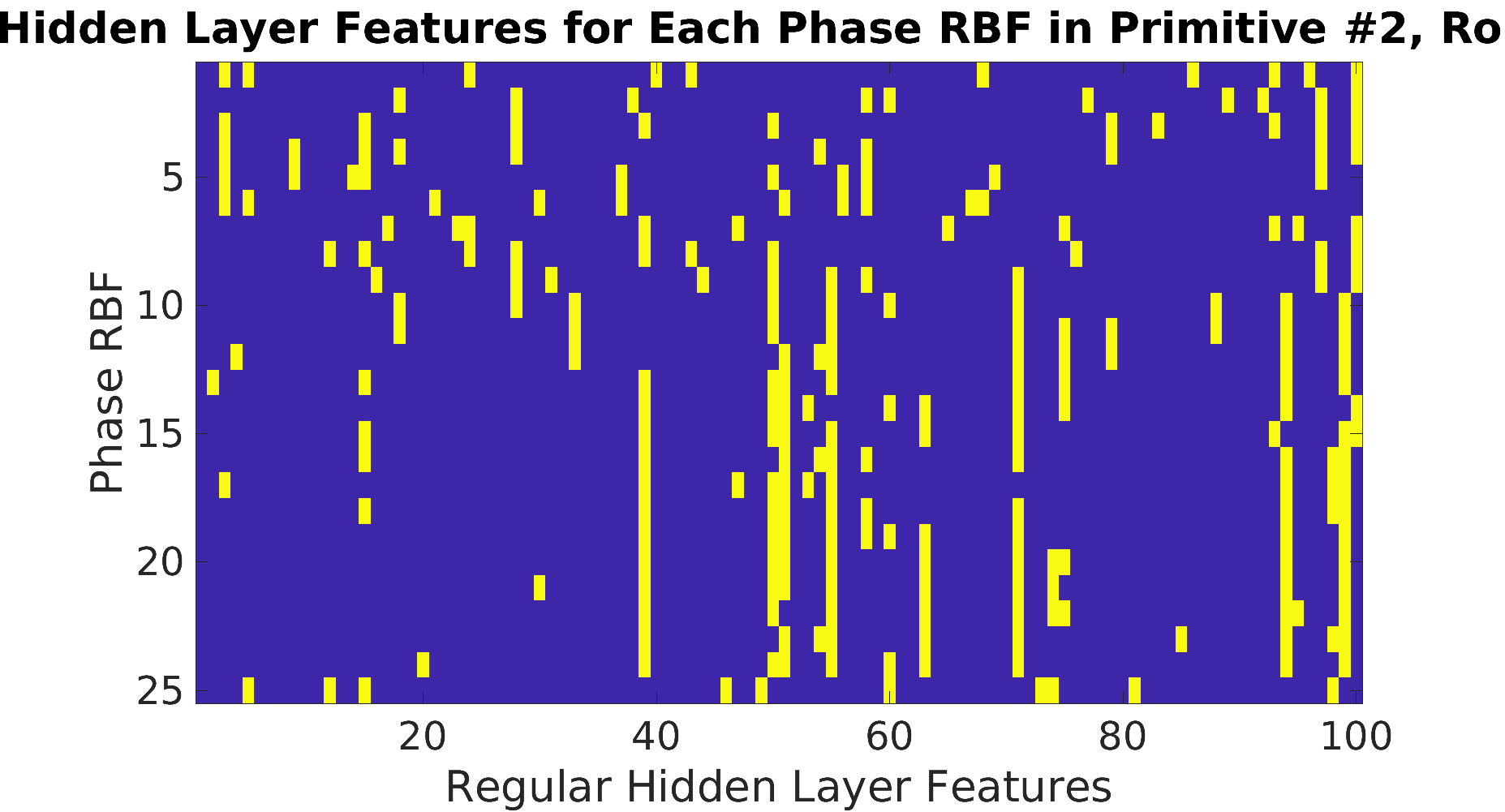}
    %
    \vspace{-0.2cm}
    \caption{(Left) comparison of regression results on primitives 2 and 3 using different neural network structures; (Middle) comparison of regression results on primitives 2 and 3 using separated feature learning (PCA or Autoencoder and phase kernel modulation) versus embedded feature learning ({\pmnn}); (Right) the top 10 dominant regular hidden layer features for each phase RBF in primitive 2, roll-orientation coupling term, displayed in yellow. 
    }
    \label{fig:nmse_and_dominant_features}
    \vspace{-0.7cm}
\end{figure*}
\item \textit{Feedback model learning}:
Now we vary the tilt stage's roll-angle to $2.5^\circ$, $5^\circ$, $7.5^\circ$, and $10^\circ$, one-at-a-time, to encode different environmental settings. At each setting, we let the robot unroll the nominal behavior. 
Beside the downward force control for contact maintenance, now we also activate the roll-orientation PI torque control at 0 Newton-meter throughout primitives 2 and 3. This allows the human demonstrator to perform the roll-orientation correction demonstration, to maintain full flat tool-tip contact relative to the now-tilted scraping board. We recorded 15 demonstrations for each setting, from which we extracted the supervised dataset for the feedback model, i.e. the pair of the sensory trace deviation $\sensortracesdeviation$ and the target coupling term $\boldsymbol{C}_{target}$ as formulated in Equation \ref{eq:OriDMPCouplingTermExtraction}. Afterwards, we learn the feedback models from this dataset using the {\pmnn}.
\item \textit{DMP and Feedback Model Unrolling/Testing}:
We test the feedback models on different settings on the robot.
\end{enumerate}
\vspace{-0.1cm}

\section{Experiments}
\label{sec:experiments}
	To evaluate the performance of the learned feedback model, we first evaluate the regression and generalization ability of the PMNNs which were trained offline on the demonstration data.
Second, we show the superiority of PMNNs over FFNNs as a choice for feedback models learning representation. Third, we investigate the importance of learning both the feature representation and the phase dependencies together within the framework of learning feedback models.
Fourth, we show the significance of the phase modulation in the feedback model learning.
Finally, we evaluate the learned feedback model's performance in making predictions of action corrections online on a real robot.

We evaluate feedback models only on primitives 2 and 3, for roll-orientation correction.
In primitive 1, we deem that there is no action correction, because the height of the dry-erase board surface is maintained constant across all settings. 

As error metric we use the normalized mean squared error (NMSE), i.e. the mean squared prediction error divided by the target coupling term's variance. 
To evaluate the learning performance of each model in our experiments, we perform a \textit{leave-one-demonstration-out} test. In this test, we perform $K$ iterations of training and testing,
where $K=15$ is the number of demonstrations per setting. At the $k$-th iteration:
\begin{itemize}
    \item The data points of the $k$-th demonstration of all settings are left-out as unseen data for \textit{generalization} testing, while the remaining $K-1$ demonstrations' data points\footnote{Each demonstration -- depending on the data collection sampling rate and demonstration duration -- provides hundreds or thousands of data points.} are shuffled randomly and split $85\%$, $7.5\%$, and $7.5\%$ for \textit{training}, \textit{validation}, and \textit{testing}, respectively.
    \item We record the training-validation-testing-generalization NMSE pairs corresponding to the lowest generalization NMSE across learning steps.
\end{itemize}
We report the mean and standard deviation of training-validation-testing-generalization NMSEs across $K$ iterations.

On all models we evaluated, we use \textit{tanh} as the activation function of the hidden layer nodes. We use the Root Mean Square Propagation (RMSProp) \cite{Tieleman2012} as the gradient descent optimization algorithm and set the \textit{dropout} \cite{Srivastava2014_NN_Dropouts} rate to 0.5.
\vspace{-0.2cm}
\subsection{Fitting and Generalization Evaluation of {\pmnn}s}
The results for primitive 2 and 3, using the {\pmnn} structure with one regular hidden layer of 100 nodes, are shown in Table \ref{tab:nmse_table}. The PMNNs achieve good training, validation, testing results, and reasonable generalization results for both primitives.
\begin{table}[ht]
\centering
\begin{tabular}{|c|c|c|c|c|}
\hline
\multirow{2}{*}{} & \multicolumn{4}{c|}{Roll-Orientation Coupling Term Learning NMSE} \\ \cline{2-5}
                  & Training & Validation & Testing & Generalization \\ \hline
Prim. 2       &0.15$\pm$0.05&0.15$\pm$0.05&0.16$\pm$0.06&0.36$\pm$0.19 \\ \hline
Prim. 3       &0.22$\pm$0.05&0.22$\pm$0.05&0.22$\pm$0.05&0.32$\pm$0.13 \\ \hline
\end{tabular}
\label{table:NMSE}
\caption{NMSE of the roll-orientation coupling term learning with \textit{leave-one-demonstration-out} test, for each primitive.}
\label{tab:nmse_table}
\vspace{-0.95cm}
\end{table}
%
\begin{figure*}[ht]
    \centering
        \null\hfill
            \subfloat{\includegraphics[trim={2.5cm 0.5cm 2.5cm 0.5cm},clip,width=0.24\textwidth]{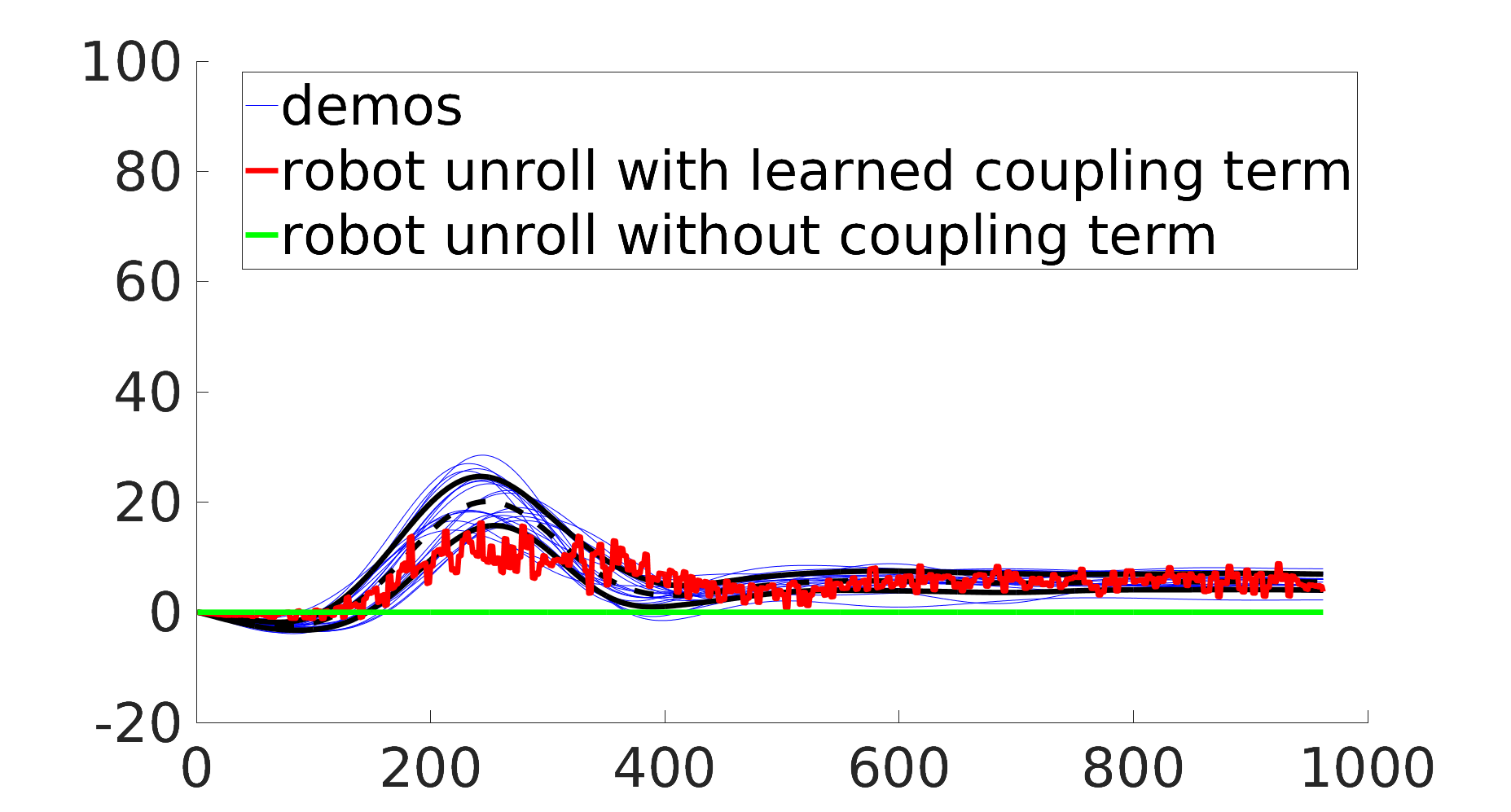}}
        \hfill
            \subfloat{\includegraphics[trim={2.5cm 0.5cm 2.5cm 0.5cm},clip,width=0.24\textwidth]{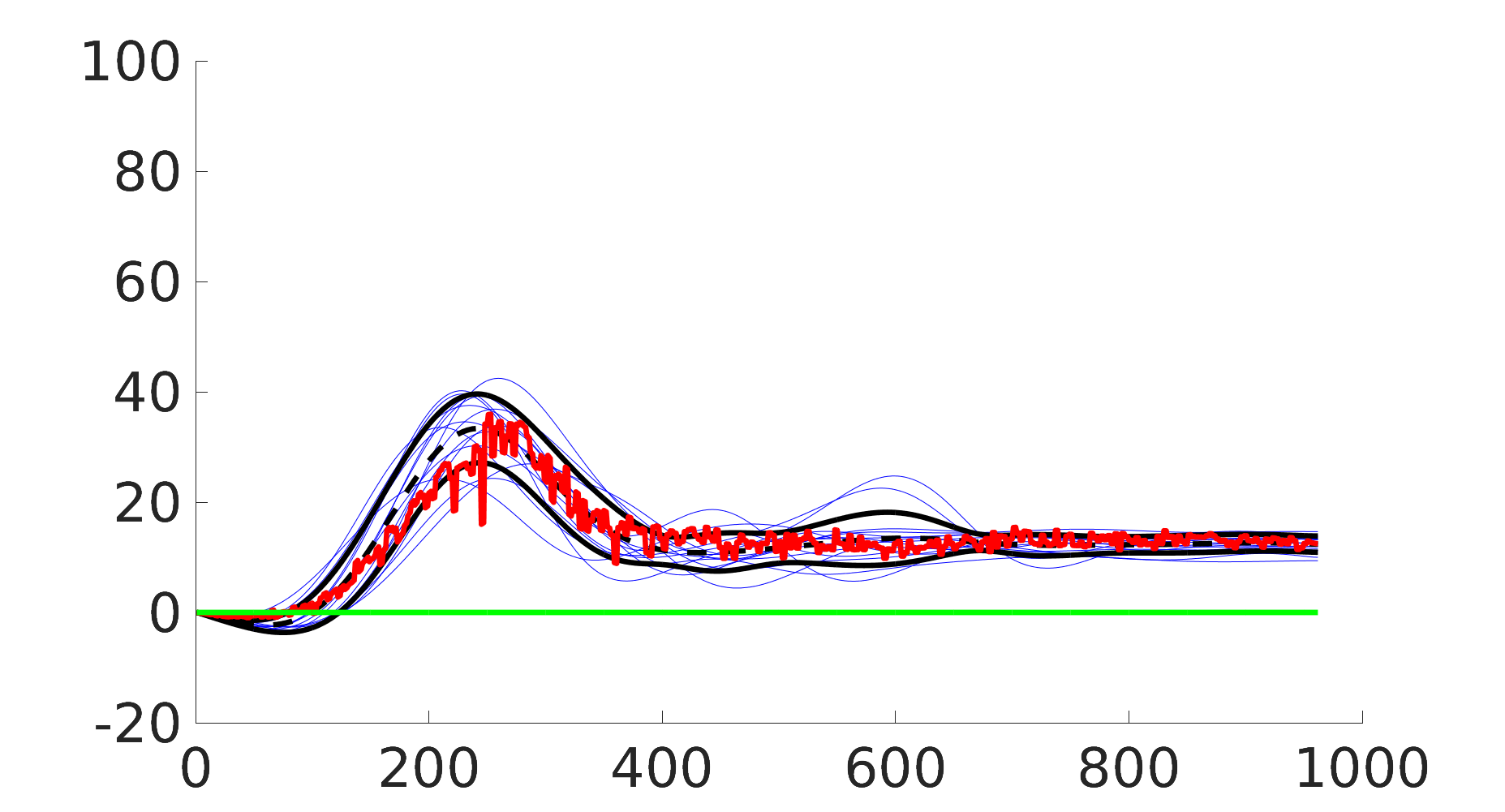}}
        \hfill
            \subfloat{\includegraphics[trim={2.5cm 0.5cm 2.5cm 0.5cm},clip,width=0.24\textwidth]{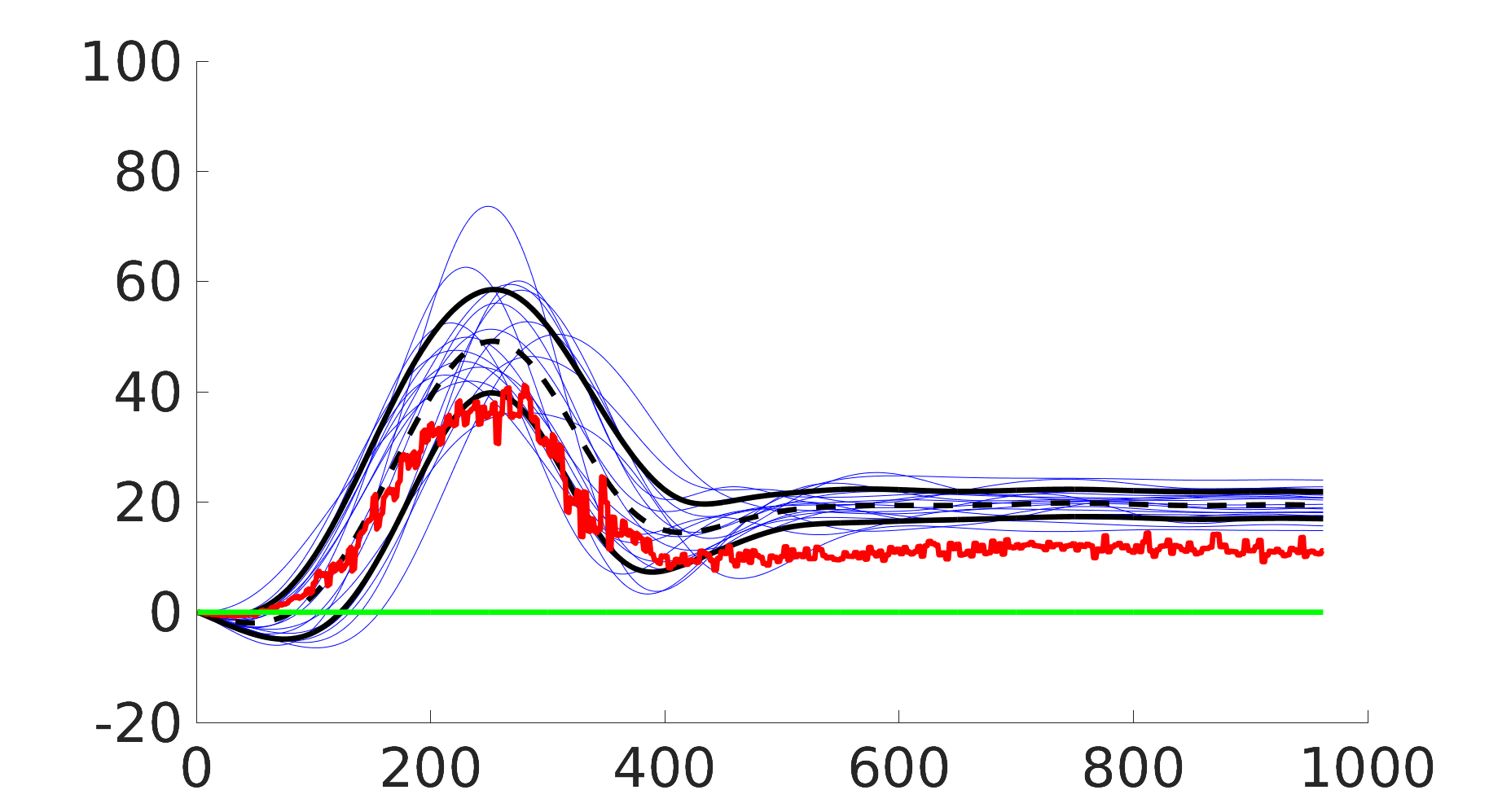}}
        \hfill
            \subfloat{\includegraphics[trim={2.5cm 0.5cm 2.5cm 0.5cm},clip,width=0.24\textwidth]{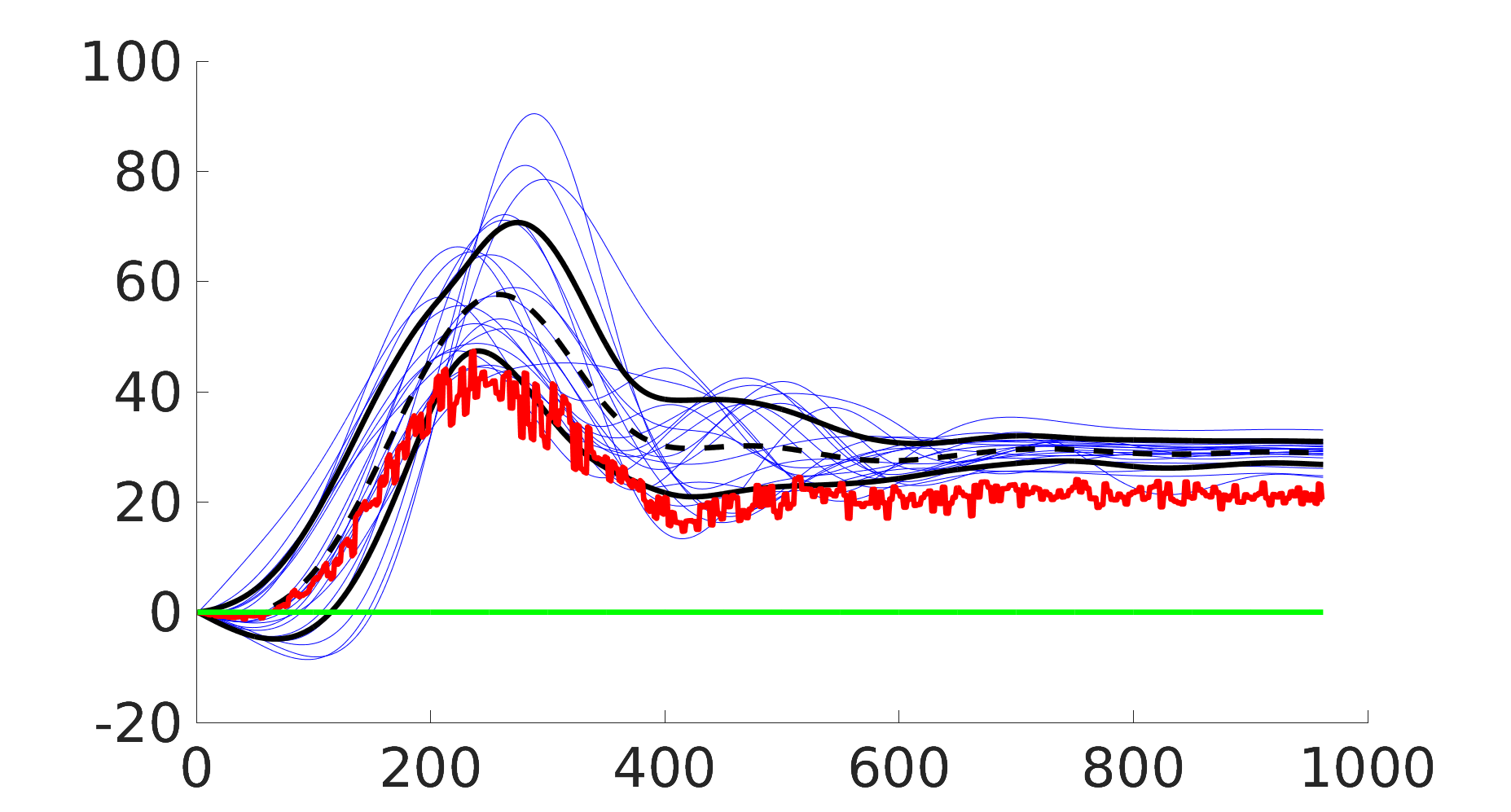}}
        \hfill\null
    \par
    \vspace{-0.7cm}
    \setcounter{subfigure}{0}
        \null\hfill
            \subfloat[Env. setting: $2.5^\circ$ roll-angle]{\includegraphics[trim={2.5cm 0.5cm 2.5cm 0.5cm},clip,width=0.24\textwidth]{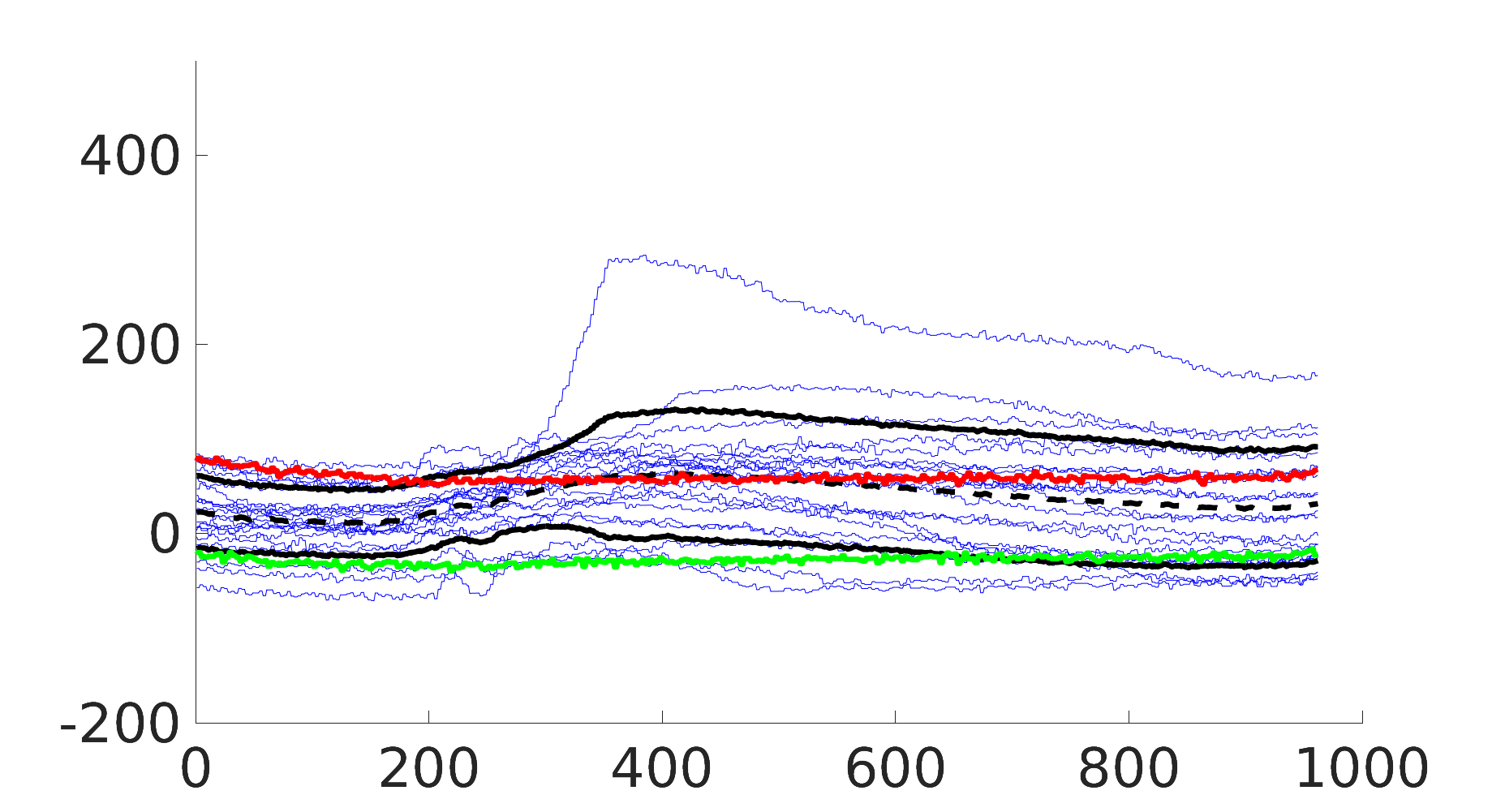}}
        \hfill
            \subfloat[Env. setting: $5.0^\circ$ roll-angle]{\includegraphics[trim={2.5cm 0.5cm 2.5cm 0.5cm},clip,width=0.24\textwidth]{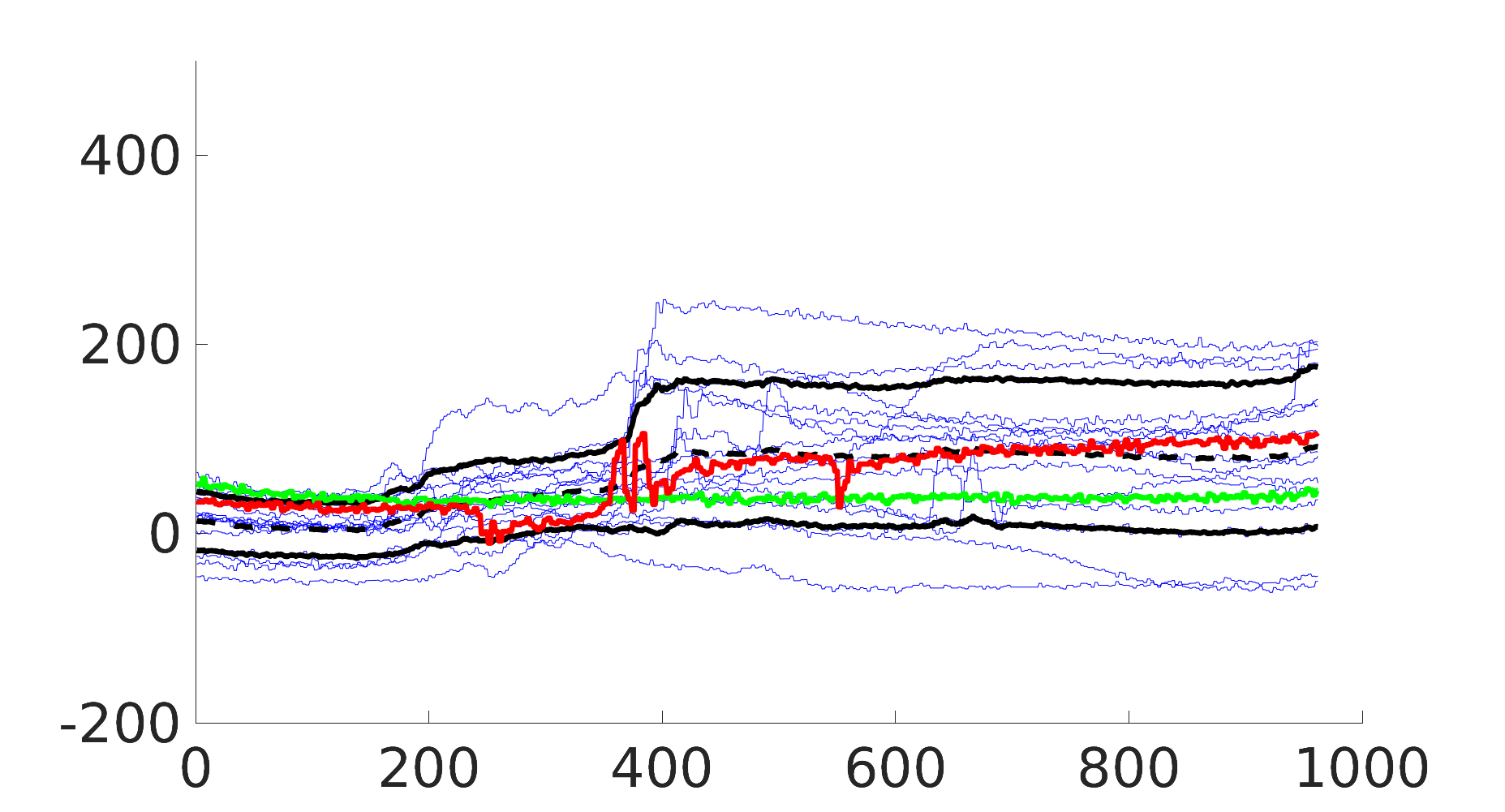}}
        \hfill
            \subfloat[Env. setting: $7.5^\circ$ roll-angle]{\includegraphics[trim={2.5cm 0.5cm 2.5cm 0.5cm},clip,width=0.24\textwidth]{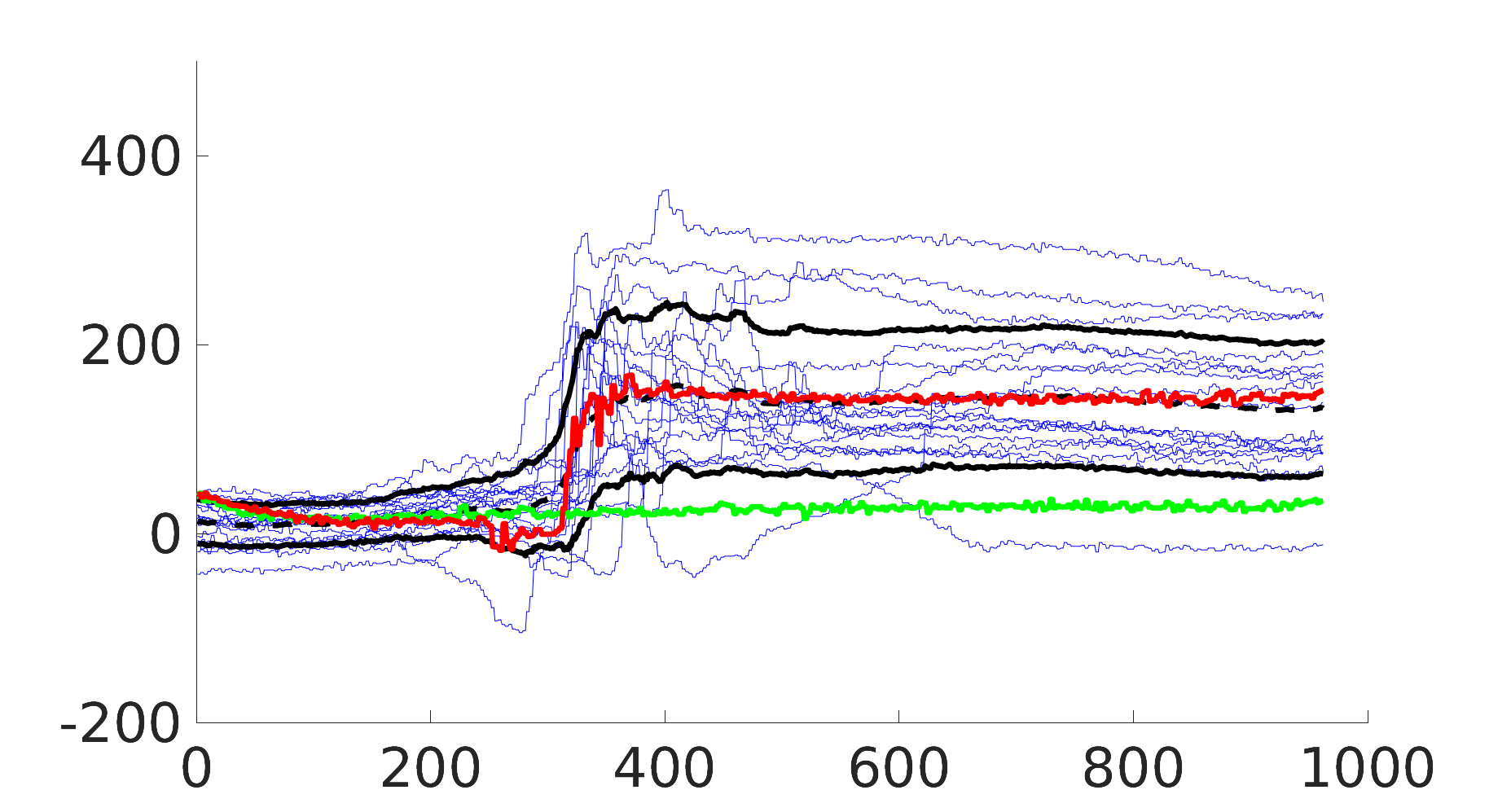}}
        \hfill
            \subfloat[Env. setting: $10.0^\circ$ roll-angle]{\includegraphics[trim={2.5cm 0.5cm 2.5cm 0.5cm},clip,width=0.24\textwidth]{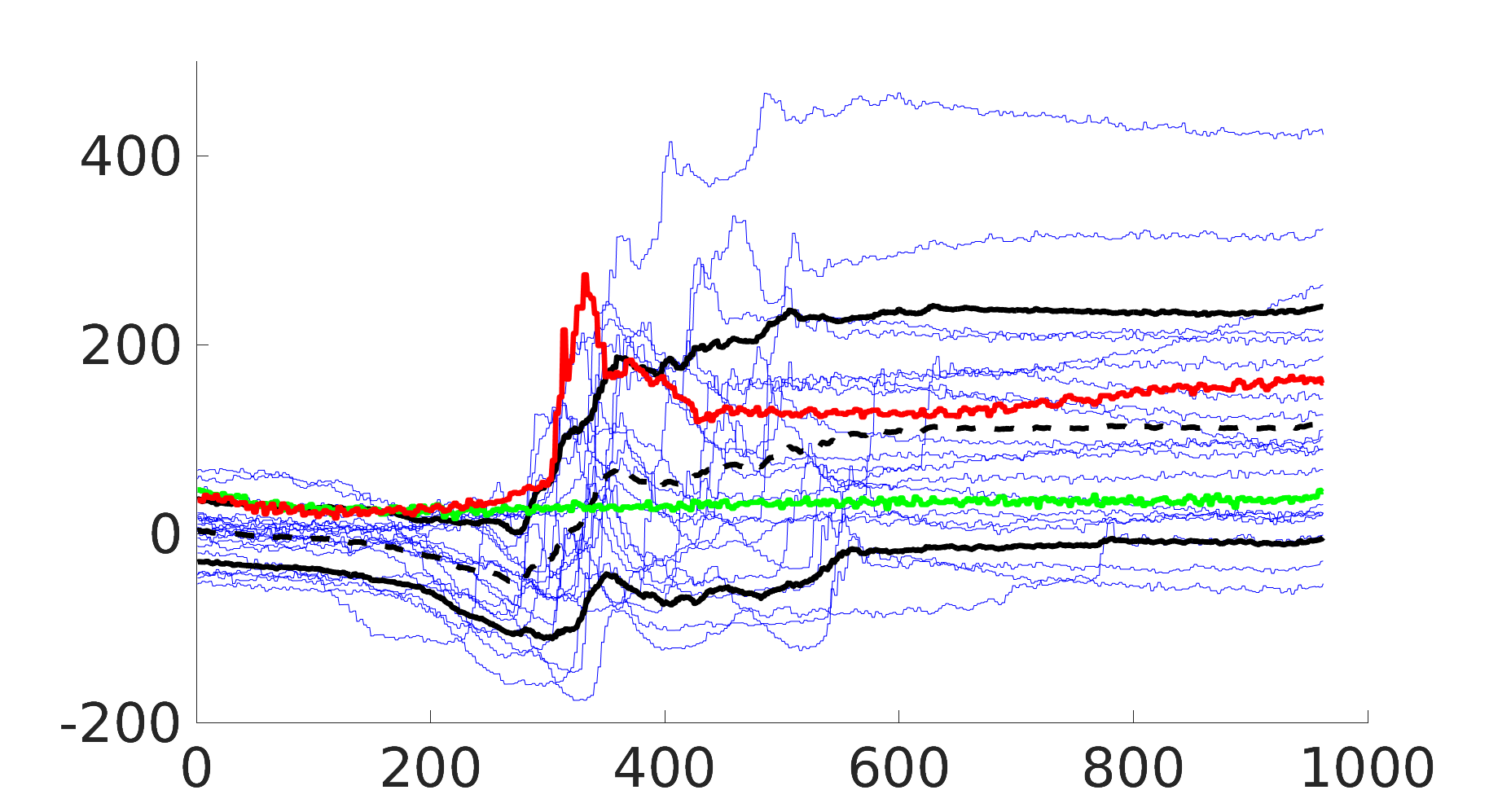}}
        \hfill\null
    \caption{The roll-orientation coupling term (top) vs. the corresponding sensor traces deviation of the right BioTac finger's electrode \#6 on primitive 2 (bottom), during scraping task on environmental (env.) setting with the tilt stage's roll-angle varies as specified in caption (a)-(d). x-axis is the time index, y-axis of top figures is the coupling term magnitude (in \textit{radians}), and y-axis of bottom figures is the discretized sensor trace deviation magnitude (unitless).}
    %
    %
    \label{fig:robot_coupling_terms_vs_sensor_traces}
    \vspace{-0.7cm}
\end{figure*}

\subsection{Performance Comparison between FFNN and {\pmnn}}
We compare the performance between FFNN and {\pmnn}. For {\pmnn}, we test two structures: one with no regular hidden layer being used, and the other with one regular hidden layer comprised of 100 nodes. For FFNN, we use two hidden layers with 100 and 25 nodes each, which is equivalent to {\pmnn} with one regular hidden layer of 100 nodes but de-activating the phase modulation. The results can be seen in Figure \ref{fig:nmse_and_dominant_features} (Left). It can be seen that {\pmnn} with one regular hidden layer of 100 nodes demonstrated the best performance compared to the other structures. 
{\pmnn} with one regular hidden layer is better than the one without regular hidden layer, most likely because of the richer learned feature representation, without getting overfitted to the data.
\vspace{-0.5cm}
\begin{figure}[t]
    \centering
    \null\hfill
        \subfloat[0.0$^{\circ}$]{\includegraphics[width=0.10125\textwidth]{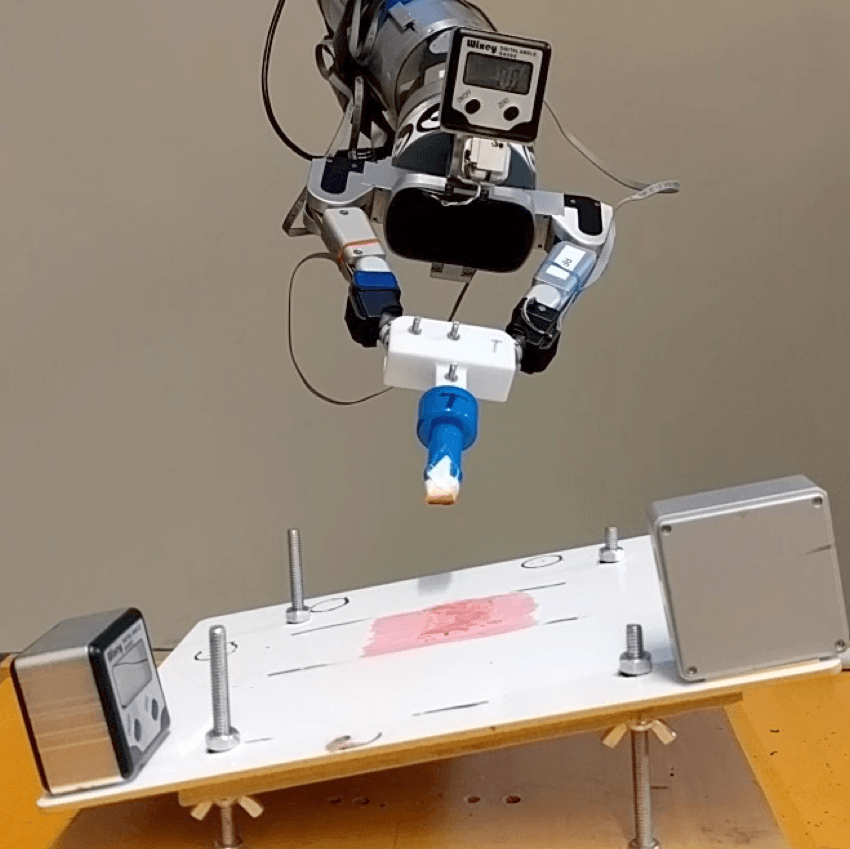}}
    \hfill
        \subfloat[0.0$^{\circ}$]{\includegraphics[width=0.10125\textwidth]{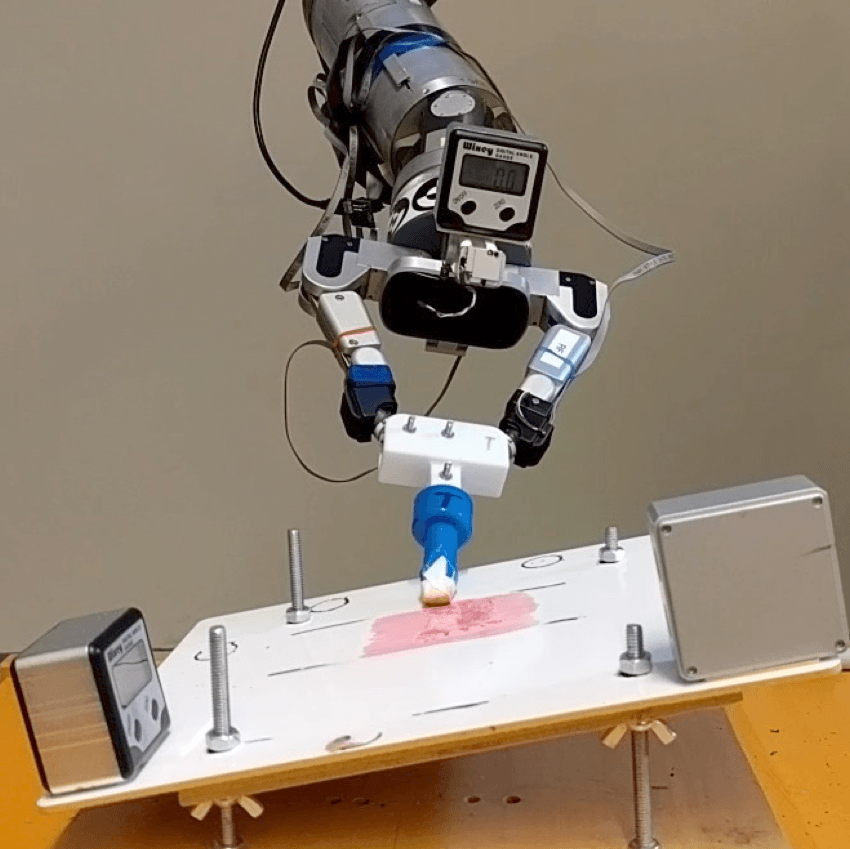}}
    \hfill
        \subfloat[0.0$^{\circ}$]{\includegraphics[width=0.10125\textwidth]{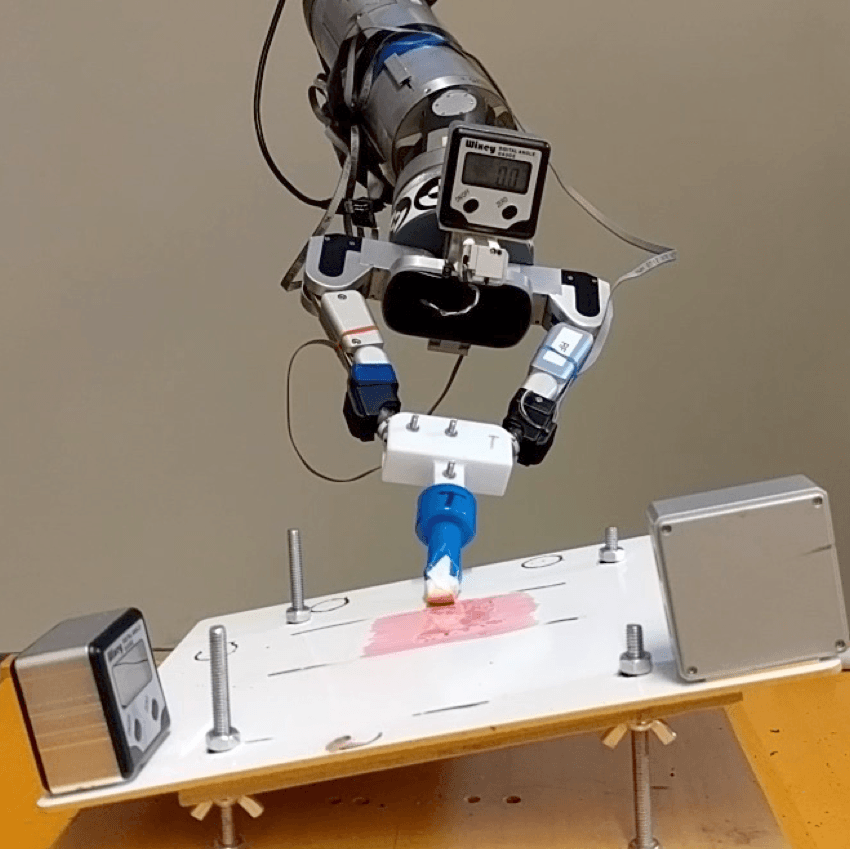}}
    \hfill
        \subfloat[2.0$^{\circ}$]{\includegraphics[width=0.10125\textwidth]{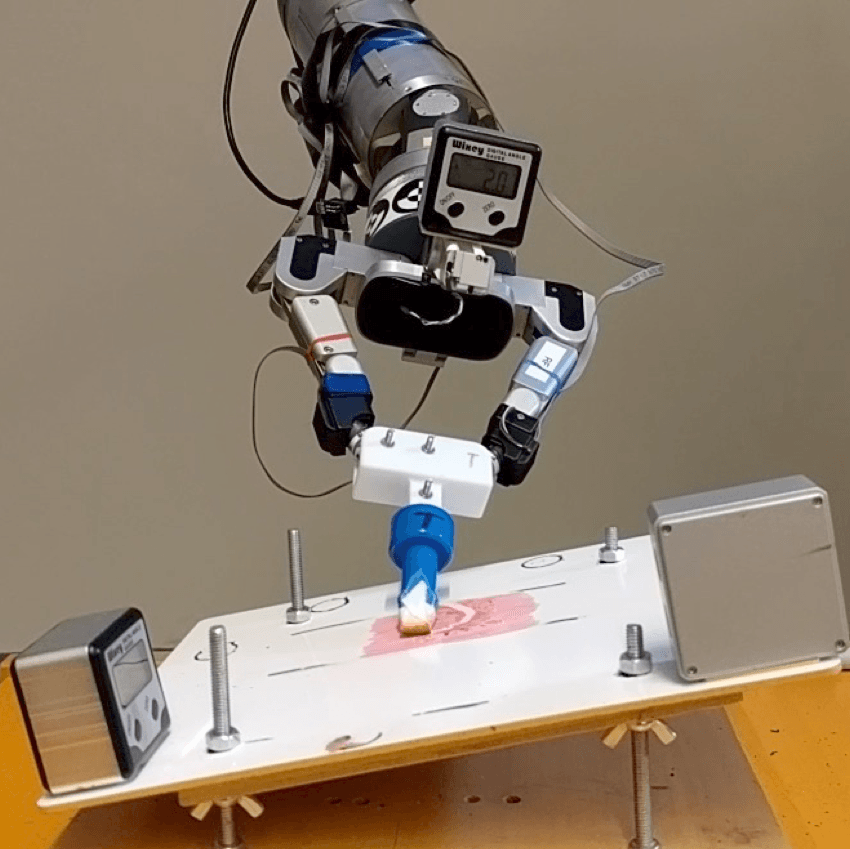}}
    \hfill\null
    \par
    \vspace{-0.2cm}
    \null\hfill
        \subfloat[0.7$^{\circ}$]{\includegraphics[width=0.10125\textwidth]{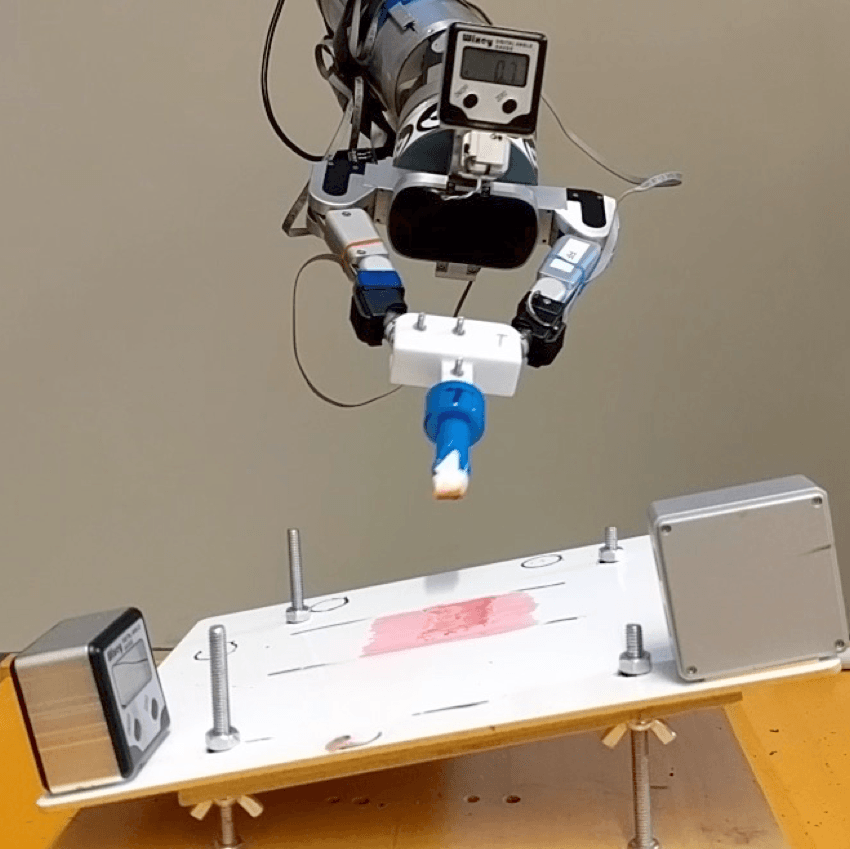}}
    \hfill
        \subfloat[2.5$^{\circ}$]{\includegraphics[width=0.10125\textwidth]{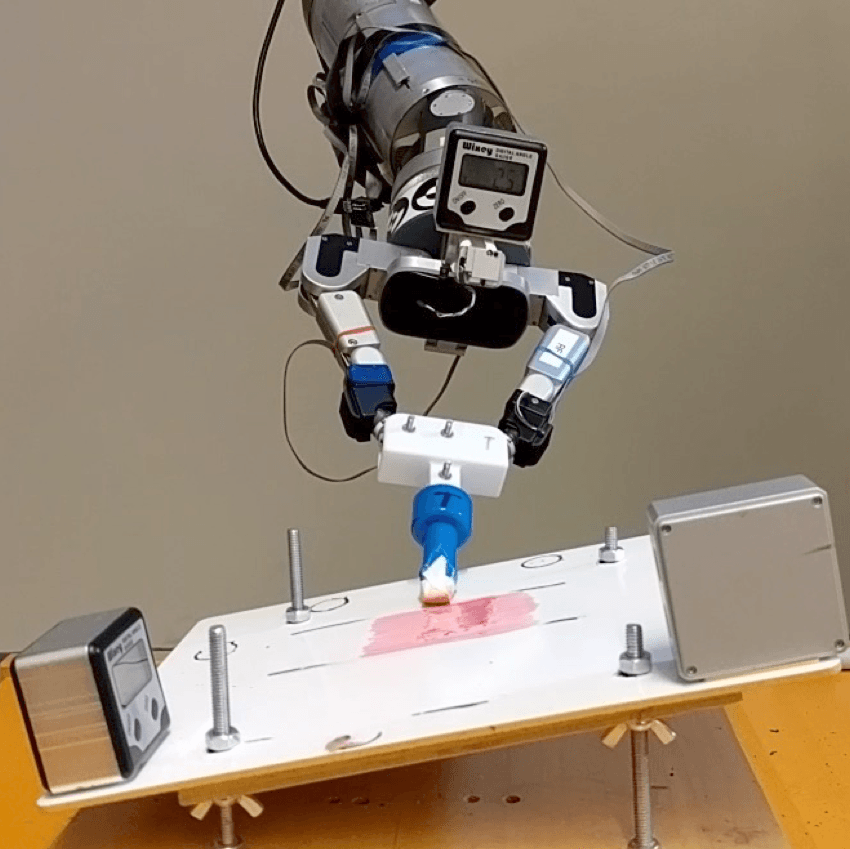}}
    \hfill
        \subfloat[5.7$^{\circ}$]{\includegraphics[width=0.10125\textwidth]{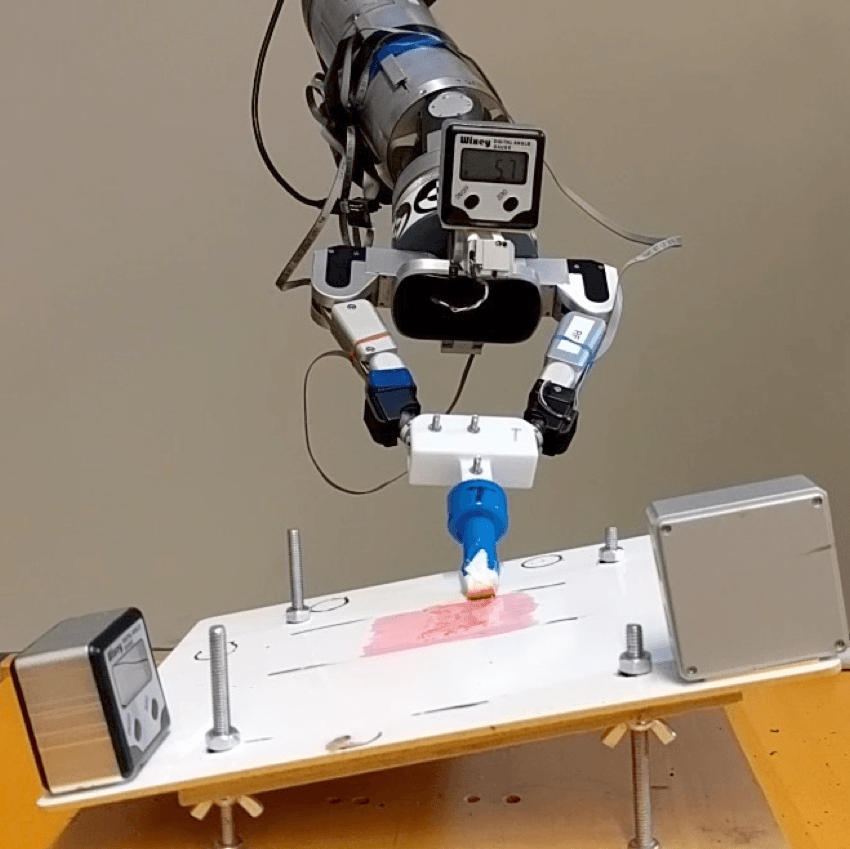}}
    \hfill
        \subfloat[3.7$^{\circ}$]{\includegraphics[width=0.10125\textwidth]{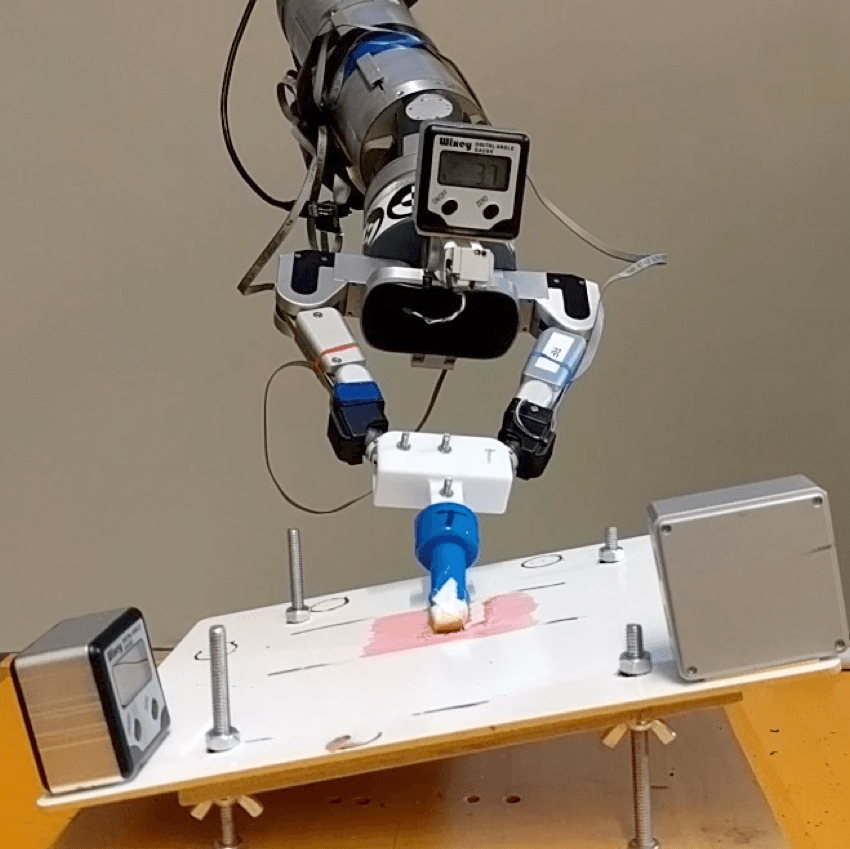}}
    \hfill\null
    \vspace{-0.15cm}
    \caption{Snapshots of our experiment on the robot while scraping on the tilt stage with +10$^\circ$ roll angle environmental setting: without adaptation (top figures, (a) to (d)) versus with adaptation (bottom figure, (e) to (h)).
    }
    \label{fig:robot_unroll_baseline_vs_coupled}
    \vspace{-0.7cm}
\end{figure}
\subsection{Comparison between Separated versus Embedded Feature Representation and Phase-Dependent Learning}
We also compare the effect of separating versus embedding the feature representation learning with overall parameter optimization under phase modulation. Chebotar et al. \cite{Chebotar_IROS_2014} used PCA for feature representation learning, which was separated from the phase-dependent parameter optimization using reinforcement learning. On the other hand, {\pmnn} embeds feature learning together with the parameter optimization under phase modulation, into an integrated process.

%
In this experiment, we used PCA which retained 99\% of the overall data variance, reducing the data dimensionality to 7 and 6 (from originally 38) for primitive 2 and 3, respectively. In addition, we also implemented an autoencoder, a non-linear dimensionality reduction method, as a substitute for PCA in representation learning. The dimensions of the latent space of the autoencoders were 7 and 6 for primitive 2 and 3, respectively. For {\pmnn}s, we used two kinds of networks: one with one regular hidden layer of 6 nodes (such that it is become comparable with the PCA counterpart), and the other with one regular hidden layer of 100 nodes.

Figure \ref{fig:nmse_and_dominant_features} (Middle) illustrates the superior performance of {\pmnn}s, due to the feature learning performed together with the phase-dependent parameter optimization. Of the two {\pmnn}s, the one with more nodes in the regular hidden layer performs better, because it can more accurately represent the mapping, while not over-fitting to the data. Based on these evaluations, we decided to use {\pmnn}s with one regular hidden layer of 100 nodes and 25 phase-modulated nodes in the final hidden layer for subsequent experiments.
\vspace{-0.1cm}
\subsection{Evaluation of Movement Phase Dependency}
Here we visualize the trained weight matrix mapping the output of 100 nodes in the regular hidden layer to the 25 nodes in the final hidden layer being modulated by the phase RBFs. This weight matrix is of dimension 25 $\times$ 100, and each row shows how each of the 100 nodes' output (or "features") in the regular hidden layer being weighted into a particular phase RBF-modulated node. In Figure \ref{fig:nmse_and_dominant_features} (Right), we display the top 10 dominant regular hidden layer node output for each phase RBF-modulated node (in yellow color), and the rest (colored in blue) are the less dominant ones. We see that between different phase RBF-modulated nodes, the priority ranking is different, suggesting that there is some dependency of the feedback on the movement phase.
\vspace{-0.2cm}
\subsection{Unrolling the Learned Feedback Model on the Robot}
In Figure \ref{fig:robot_unroll_baseline_vs_coupled}, we show the snapshots of our robot scraping experiment on a setting with $10^\circ$ roll-angle of the tilt stage. In particular, we compare between the nominal plan execution (top figures, from (a) to (d)) and the adapted plan execution (bottom figures, from (e) to (h), using the trained feedback models).
From left to right ((a) to (d), and (e) to (h)), it shows subsequent phases of plan execution. The caption ((a) to (h)) shows the reading of the Digital Angle Gauge mounted on top of the middle finger of the hand.
We see that if we turn off the coupling term (nominal plan execution, top figures), there was no correction applied to the tool-tip orientation and the scraping result was worse than when the online adaptation was applied (adapted plan execution, bottom figures).

Figure \ref{fig:robot_coupling_terms_vs_sensor_traces} shows the coupling term (top) alongside the corresponding sensor trace deviation of one of the electrodes (bottom) during plan execution at 4 different environmental settings as specified in caption (a)-(d). We compare between several cases: human demonstrations (blue), human demonstrations' mean trajectory (dashed black), range of demonstrations within 1 standard deviation from the mean trajectory (solid black), during robot unrolling of the nominal behavior (green), and during robot unrolling while applying the coupling term computed online by the trained feedback model (red). On the top plots, we see that the trained feedback model can differentiate between settings and apply the approximately correct amount of correction. When applying the coupling term computed online by the trained feedback model, the sensor trace deviation is also close to those of demonstrations, as shown in the bottom plots.

Finally, video \url{https://youtu.be/7Dx5imy1Kcw} shows the scraping execution at two settings, at $5^\circ$ and $10^\circ$ roll-angle of the tilt stage, while applying the corrections predicted online by the trained feedback model.

\section{Conclusion}
\label{sec:discussion}
	We introduced a general framework for learning-from-demonstration of feedback models, mapping sensory trace deviations to action corrections. In particular, we introduced phase-modulated neural networks ({\pmnn}s), which allow to fit phase-dependent feedback models and preserve the convergence properties of DMPs. Finally, we demonstrate the superior learning performance of our {\pmnn}-based framework when compared to state-of-the-art methods, as well as its capability in performing online adaptation on a real robot.




\section*{APPENDIX}
\label{sec:appendix}
Unit quaternion is a hypercomplex number which can be written as a vector $\boldsymbol{Q} = 
\begin{bmatrix} 
	r & \boldsymbol{q}^T 
\end{bmatrix}^T$, such that $\|\boldsymbol{Q}\| = 1$ with $r$ and $\boldsymbol{q} = \begin{bmatrix} 
	q_1 & q_2 & q_3
\end{bmatrix}^T$ are the real scalar and the vector of three imaginary components of the quaternions, respectively.
For computation with orientation trajectory, several operations needs to be defined as follows:
\begin{itemize}
	\item quaternion composition operation:
        \begin{equation}
        	\boldsymbol{Q}_A \circ \boldsymbol{Q}_B = 
        \begin{bmatrix} 
        	r_A	& -q_{A1}	& -q_{A2}	& -q_{A3}	\\
        	q_{A1}	& r_A		& -q_{A3	}	& q_{A2}		\\
        	q_{A2}	& q_{A3}		& r_A		& -q_{A1}	\\
        	q_{A3}	& -q_{A2}	& q_{A1}		& r_A
        \end{bmatrix}
        \begin{bmatrix} 
        	r_B	\\
        	q_{B1} 	\\
        	q_{B2}	\\
        	q_{B3}
        \end{bmatrix}
        \label{eq:QuatComposition}
        \end{equation}
	\item quaternion conjugation operation:
        \begin{equation}
        \boldsymbol{Q}^{*} = 
        \begin{bmatrix} 
        	r \\
        	-\boldsymbol{q} 
        \end{bmatrix}
        \label{eq:QuatConjugation}
        \end{equation}
    \item \textit{logarithm mapping} ($\log(\cdot)$ operation), which maps an element of $\textit{SO}$(3) to $\textit{so}$(3), is defined as:
        \begin{equation}
        	\log\left( \boldsymbol{Q} \right) = \log\left( 
        	\begin{bmatrix} 
        		r \\
        		\boldsymbol{q} 
        	\end{bmatrix}	
        	\right) = 
        	\frac{\arccos{(r)}}{\sin{(\arccos{(r)})}} \boldsymbol{q}
        	\label{eq:LogMapping}
        \end{equation}
    \item \textit{exponential mapping} ($\exp(\cdot)$ operation, the inverse of $\log(\cdot)$ operation) maps an element of $\textit{so}$(3) to $\textit{SO}$(3):
        \begin{equation}
        	\exp\left( \boldsymbol{\omega} \right) = 
        	    \begin{bmatrix} 
        			\cos{\left( \| \boldsymbol{\omega} \| \right)} \\
        			\frac{\sin{\left( \| \boldsymbol{\omega} \| \right)}}{\| \boldsymbol{\omega} \|} \boldsymbol{\omega}
        	    \end{bmatrix}
        	\label{eq:ExpMapping}
        \end{equation}
\end{itemize}

\section*{ACKNOWLEDGMENT}
We thanked Gerald E. Loeb for the support on BioTac sensors, Oliver Kroemer for the scraping task testbed suggestion, as well as Ludovic Righetti, Vincent Enachescu, and Ryan Julian for reviewing initial drafts of the paper.


\bibliographystyle{IEEEtran}
\bibliography{references}

\begin{thebibliography}{10}
\providecommand{\url}[1]{#1}
\csname url@rmstyle\endcsname
\providecommand{\newblock}{\relax}
\providecommand{\bibinfo}[2]{#2}
\providecommand\BIBentrySTDinterwordspacing{\spaceskip=0pt\relax}
\providecommand\BIBentryALTinterwordstretchfactor{4}
\providecommand\BIBentryALTinterwordspacing{\spaceskip=\fontdimen2\font plus
\BIBentryALTinterwordstretchfactor\fontdimen3\font minus
  \fontdimen4\font\relax}
\providecommand\BIBforeignlanguage[2]{{%
\expandafter\ifx\csname l@#1\endcsname\relax
\typeout{** WARNING: IEEEtran.bst: No hyphenation pattern has been}%
\typeout{** loaded for the language `#1'. Using the pattern for}%
\typeout{** the default language instead.}%
\else
\language=\csname l@#1\endcsname
\fi
#2}}

\bibitem{pastor_IROS_2011_ASM}
P.~Pastor, L.~Righetti, M.~Kalakrishnan, and S.~Schaal, ``Online movement
  adaptation based on previous sensor experiences,'' in \emph{IEEE/RSJ
  International Conference on Intelligent Robots and Systems}, 2011, pp.
  365--371.

\bibitem{pastor2013dynamic}
P.~Pastor, M.~Kalakrishnan, F.~Meier, F.~Stulp, J.~Buchli, E.~Theodorou, and
  S.~Schaal, ``From dynamic movement primitives to associative skill
  memories,'' \emph{Robotics and Autonomous Systems}, vol.~61, no.~4, pp.
  351--361, 2013.

\bibitem{rai2014learning}
A.~Rai, F.~Meier, A.~Ijspeert, and S.~Schaal, ``Learning coupling terms for
  obstacle avoidance,'' in \emph{IEEE-RAS International Conference on Humanoid
  Robots}, 2014, pp. 512--518.

\bibitem{icra2017_learning_feedback}
A.~Rai, G.~Sutanto, S.~Schaal, and F.~Meier, ``Learning feedback terms for
  reactive planning and control,'' in \emph{Proceedings of the IEEE
  International Conference on Robotics and Automation (ICRA)}, 2017.

\bibitem{Chebotar_IROS_2014}
Y.~Chebotar, O.~Kroemer, and J.~Peters, ``Learning robot tactile sensing for
  object manipulation,'' in \emph{IEEE/RSJ International Conference on
  Intelligent Robots and Systems}, 2014, pp. 3368--3375.

\bibitem{Ijspeert_NC_2013}
A.~J. Ijspeert, J.~Nakanishi, H.~Hoffmann, P.~Pastor, and S.~Schaal,
  ``Dynamical movement primitives: Learning attractor models for motor
  behaviors,'' \emph{Neural Comput.}, vol.~25, no.~2, pp. 328--373, 2013.

\bibitem{Kramberger_Humanoids16_GenOrientationDMP}
A.~Kramberger, A.~Gams, B.~Nemec, and A.~Ude, ``Generalization of orientational
  motion in unit quaternion space,'' in \emph{IEEE-RAS International Conference
  on Humanoid Robots}, 2016, pp. 808--813.

\bibitem{Ude_ICRA14_OrientationDMP}
A.~Ude, B.~Nemec, T.~Petric, and J.~Morimoto, ``Orientation in cartesian space
  dynamic movement primitives,'' in \emph{IEEE International Conference on
  Robotics and Automation}, 2014, pp. 2997--3004.

\bibitem{nemec2012action}
B.~Nemec and A.~Ude, ``Action sequencing using dynamic movement primitives,''
  \emph{Robotica}, vol.~30, no.~05, pp. 837--846, 2012.

\bibitem{Park_Humanoids_2008}
D.-H. Park, H.~Hoffmann, P.~Pastor, and S.~Schaal, ``Movement reproduction and
  obstacle avoidance with dynamic movement primitives and potential fields,''
  in \emph{IEEE International Conference on Humanoid Robots}, 2008, pp. 91--98.

\bibitem{Hoffmann_ICRA_2009}
H.~Hoffmann, P.~Pastor, D.~H. Park, and S.~Schaal, ``Biologically-inspired
  dynamical systems for movement generation: Automatic real-time goal
  adaptation and obstacle avoidance,'' in \emph{IEEE International Conference
  on Robotics and Automation}, 2009, pp. 2587--2592.

\bibitem{Khansari_IJRR_2016}
M.~Khansari, E.~Klingbeil, and O.~Khatib, ``Adaptive human-inspired compliant
  contact primitives to perform surface–surface contact under uncertainty,''
  \emph{The International Journal of Robotics Research}, vol.~35, no.~13, pp.
  1651--1675, 2016.

\bibitem{hoffmann2014adaptive}
H.~Hoffmann, Z.~Chen, D.~Earl, D.~Mitchell, B.~Salemi, and J.~Sinapov,
  ``Adaptive robotic tool use under variable grasps,'' \emph{Robotics and
  Autonomous Systems}, vol.~62, no.~6, pp. 833--846, 2014.

\bibitem{Abu_Dakka2015}
F.~J. Abu-Dakka, B.~Nemec, J.~A. J{\o}rgensen, T.~R. Savarimuthu,
  N.~Kr{\"u}ger, and A.~Ude, ``Adaptation of manipulation skills in physical
  contact with the environment to reference force profiles,'' \emph{Autonomous
  Robots}, vol.~39, no.~2, pp. 199--217, Aug 2015.

\bibitem{gams2015learning}
A.~Gams, M.~Denisa, and A.~Ude, ``Learning of parametric coupling terms for
  robot-environment interaction,'' in \emph{IEEE International Conference on
  Humanoid Robots}, 2015, pp. 304--309.

\bibitem{Kupcsiketal15}
A.~Kupcsik, M.~Deisenroth, J.~Peters, L.~Ai~Poh, V.~Vadakkepat, and G.~Neumann,
  ``Model-based contextual policy search for data-efficient generalization of
  robot skills,'' \emph{Artificial Intelligence}, vol. 247, pp. 415--439, 2017.

\bibitem{Sung_ICRA_2017}
J.~Sung, J.~K. Salisbury, and A.~Saxena, ``Learning to represent haptic
  feedback for partially-observable tasks,'' in \emph{IEEE International
  Conference on Robotics and Automation}, 2017, pp. 2802--2809.

\bibitem{Kober_IROS_2008}
J.~Kober, B.~Mohler, and J.~Peters, ``Learning perceptual coupling for motor
  primitives,'' in \emph{IEEE/RSJ International Conference on Intelligent
  Robots and Systems}, 2008, pp. 834--839.

\bibitem{Bishop_1991_RBFNN}
C.~Bishop, ``Improving the generalization properties of radial basis function
  neural networks,'' \emph{Neural Computation}, vol.~3, no.~4, pp. 579--588,
  1991.

\bibitem{TensorFlowBib}
\BIBentryALTinterwordspacing
M.~Abadi, A.~Agarwal, P.~Barham, E.~Brevdo, Z.~Chen, C.~Citro, G.~Corrado,
  A.~Davis, J.~Dean, M.~Devin, S.~Ghemawat, I.~Goodfellow, A.~Harp, G.~Irving,
  M.~Isard, Y.~Jia, R.~Jozefowicz, L.~Kaiser, M.~Kudlur, J.~Levenberg,
  D.~Mané, R.~Monga, S.~Moore, D.~Murray, C.~Olah, M.~Schuster, J.~Shlens,
  B.~Steiner, I.~Sutskever, K.~Talwar, P.~Tucker, V.~Vanhoucke, V.~Vasudevan,
  F.~Viégas, O.~Vinyals, P.~Warden, M.~Wattenberg, M.~Wicke, Y.~Yu, and
  X.~Zheng, ``Tensorflow: Large-scale machine learning on heterogeneous
  distributed systems,'' 2015. [Online]. Available:
  \url{http://download.tensorflow.org/paper/whitepaper2015.pdf}
\BIBentrySTDinterwordspacing

\bibitem{Srivastava2014_NN_Dropouts}
N.~Srivastava, G.~Hinton, A.~Krizhevsky, I.~Sutskever, and R.~Salakhutdinov,
  ``Dropout: A simple way to prevent neural networks from overfitting,''
  \emph{J. Mach. Learn. Res.}, vol.~15, no.~1, pp. 1929--1958, 2014.

\bibitem{wettels08}
N.~Wettels, V.~Santos, R.~Johansson, and G.~Loeb, ``Biomimetic tactile sensor
  array.'' \emph{Advanced Robotics}, vol.~22, no.~8, pp. 829--849, 2008.

\bibitem{fod2002automated}
A.~Fod, M.~J. Matari{\'c}, and O.~C. Jenkins, ``Automated derivation of
  primitives for movement classification,'' \emph{Autonomous robots}, vol.~12,
  no.~1, pp. 39--54, 2002.

\bibitem{Tieleman2012}
T.~Tieleman and G.~Hinton, ``{Lecture 6.5---RmsProp: Divide the gradient by a
  running average of its recent magnitude},'' COURSERA: Neural Networks for
  Machine Learning, 2012.

\end{thebibliography}

\end{document}